\documentclass[lettersize,journal]{IEEEtran}
\usepackage{amsmath,amsfonts}
\usepackage{algorithmic}
\usepackage{array}
\usepackage[caption=false,font=normalsize,labelfont=sf,textfont=sf]{subfig}
\usepackage{textcomp}
\usepackage{stfloats}
\usepackage{url}
\usepackage{verbatim}
\usepackage{graphicx}
\usepackage{cite}
\usepackage[misc]{ifsym}
\usepackage[pagebackref=false,breaklinks=true,colorlinks=true,bookmarks=false]{hyperref}
\usepackage{epsfig}
\usepackage{amssymb}
\usepackage{enumitem}
\usepackage{makecell}
\usepackage{cases}
\usepackage{xcolor}
\usepackage{multirow}
\usepackage[normalem]{ulem}
\usepackage[ruled,vlined,linesnumbered]{algorithm2e}
\usepackage{booktabs}
\usepackage{pifont}

\makeatletter
\DeclareRobustCommand\onedot{\futurelet\@let@token\@onedot}
\def\@onedot{\ifx\@let@token.\else.\null\fi\xspace}

\usepackage{xcolor}
\usepackage{soul}      
\sethlcolor{yellow}       

\soulregister\cite7
\soulregister\ref7
\soulregister\eqref7

\makeatother
\usepackage{amssymb}
\usepackage{bbding}
\def\methodname{TriMM}
\begin{document}

\title{Collaborative Multi-Modal Coding for
High-Quality 3D Generation}

\author{
    Ziang~Cao,
    Zhaoxi~Chen,
    Liang~Pan\textsuperscript{\Letter},
    Ziwei~Liu\textsuperscript{\Letter}
    \thanks{
        \IEEEcompsocthanksitem Ziang Cao, Zhaoxi Chen, Ziwei Liu are with
        S-Lab, Nanyang Technological University, Singapore.
        \textit{\{ziang001, zhaoxi001\}@e.ntu.edu.sg, \{ziwei.liu \}@ntu.edu.sg}
        \IEEEcompsocthanksitem Liang Pan is with
        Shanghai Artificial Intelligence Laboratory
        \textit{panliang1@pjlab.org.cn}
        \IEEEcompsocthanksitem \textsuperscript{\Letter} Corresponding Author
    }
}

\markboth{IEEE TRANSACTIONS ON PATTERN ANALYSIS AND MACHINE INTELLIGENCE, VOL. X, NO. X, MMMMMMM YYYY}%
{CAO \MakeLowercase{\textit{et al.}}: Collaborative Multi-Modal Coding for High-Quality 3D Generation}

\maketitle

\begin{abstract}
    3D content inherently encompasses multi-modal characteristics and can be projected into different modalities (\textit{e.g.}, RGB images, RGBD, and point clouds). Each modality exhibits distinct advantages in 3D asset modeling: RGB images contain vivid 3D textures, whereas point clouds define fine-grained 3D geometries. However, most existing 3D-native generative architectures either operate predominantly within single-modality paradigms—thus overlooking the complementary benefits of multi-modality data—or restrict themselves to 3D structures, thereby limiting the scope of available training datasets. To holistically harness multi-modalities for 3D modeling, we present \textbf{\methodname}, the first feed-forward 3D-native generative model that learns from basic multi-modalities (\textit{e.g.}, RGB, RGBD, and point cloud). Specifically, \textbf{1)} \methodname~first introduces collaborative multi-modal coding, which integrates modality-specific features while preserving their unique representational strengths. \textbf{2)} Furthermore, auxiliary 2D and 3D supervision are introduced to raise the robustness and performance of multi-modal coding. 
\textbf{3)} Based on the embedded multi-modal code, \methodname\ employs a triplane latent diffusion model to generate 3D assets of superior quality, enhancing both the texture and the geometric detail. Extensive experiments on multiple well-known datasets demonstrate that \methodname, by effectively leveraging multi-modality, achieves competitive performance with models trained on large-scale datasets, despite utilizing a small amount of training data. Furthermore, we conduct additional experiments on recent RGB-D datasets, verifying the feasibility of incorporating other multi-modal datasets into 3D generation.
\end{abstract}

\section{Introduction}
\IEEEPARstart{T}{h}e automatic generation of high-quality 3D assets from images or texts has a wide range of applications, from virtual reality and robotics simulation to industrial design and animation.
Recent advances have been witnessed in image~\cite{rombach2022high} and video generation~\cite{blattmann2023stable} given open-source billion-scale 2D datasets~\cite{schuhmann2022laion}. 
However, the largest public 3D dataset, Objaverse~\cite{deitke2023objaverse,deitke2024objaverse} only includes millions of 3D objects.
Therefore, to compensate for the data-hungry of 3D generative models, effectively leveraging heterogeneous 3D datasets through methodological designs is crucial for scalable 3D generative models.

\begin{figure*}[t]
	\centering

	\includegraphics[width=1\textwidth]{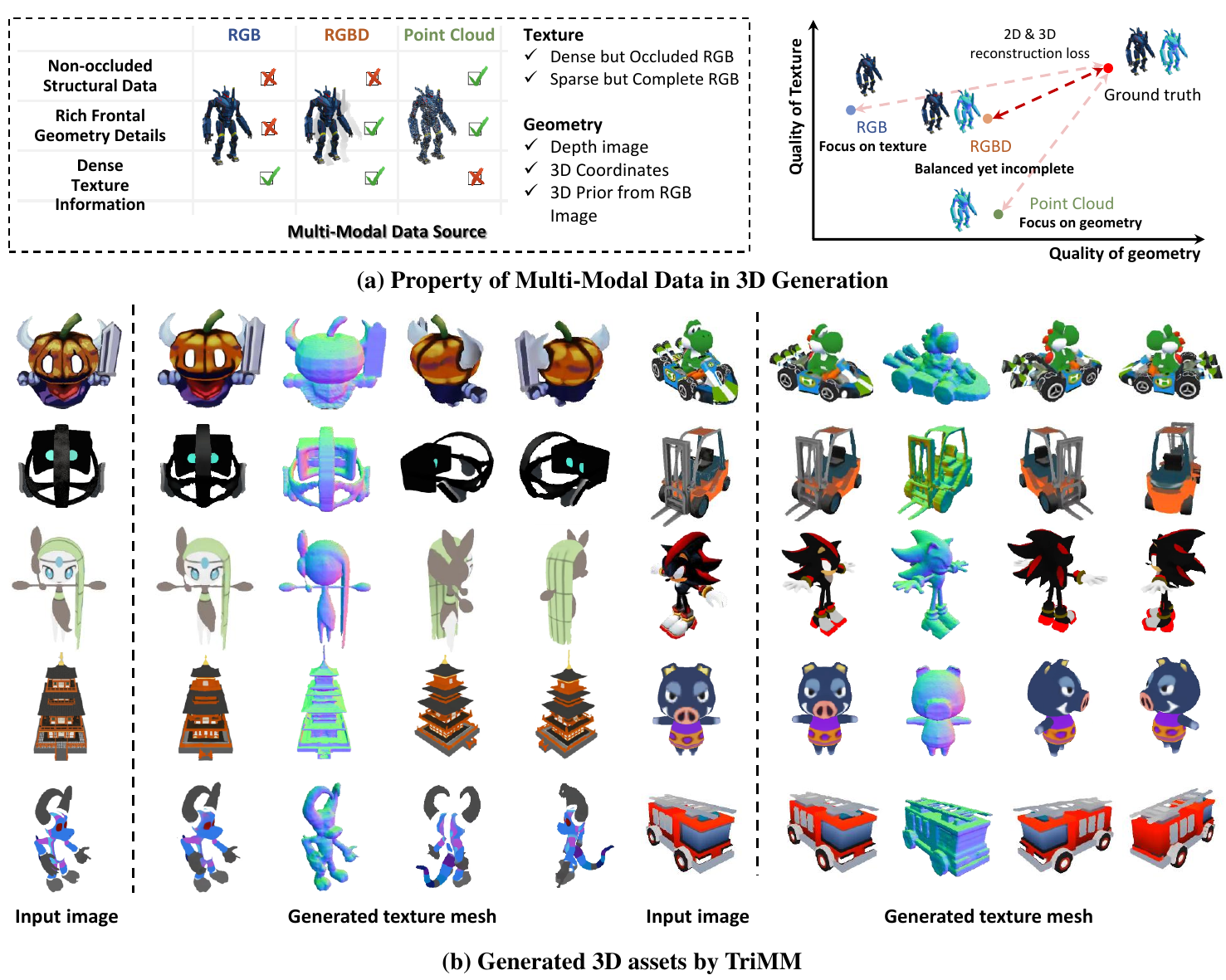}
	\vspace{-12pt}
	\caption{By leveraging \textbf{a) Collaborative Multi-Modal Coding} encoded from photometric (RGB, RGBD) and geometric (RGBD, Point Clouds) information, \textbf{b) \methodname\ }can create high-quality textured meshes within 4 seconds from a single image.}
	\vspace{-10pt}
	\label{fig:teaser}
	
\end{figure*}

Prior arts on 3D generation could be further categorized into three: 1) optimization-based methods~\cite{poole2022dreamfusion} using Score Distillation Sampling (SDS); 2) reconstruction-based feedforward models~\cite{lrm, tang2025lgm, instantmesh} which employ a transformer to regress 3D representation given sparse-view images; 3) 3D-native feedforward models~\cite{3dtopia,chen20243dtopia,muller2023diffrf,cao2023large,cao2024difftf++,lan2024ga,chen2024sar3d,lan2024ln3diff,luo20243denhancer} with 3D diffusion or autoregressive framework. Current 3D generative approaches, despite architectural variations, predominantly adhere to monolithic modality paradigms—typically relying on colored renderings as main training sources. While RGB images provide dense texture priors of 3D assets (\textit{e.g.}, material reflectance, specular highlights), this photometric modality suffers from fundamental limitations: 1) geometric ambiguity in occluded regions, and 2) topological uncertainty due to projective viewpoint. 
These modality-specific constraints limit the scalability and quality of existing 3D generative models.

To address this challenge, we propose \methodname, a collaborative multi-modal coding for high-quality feed-forward 3D generation. 
At the core of our framework is a unified latent coding that synergistically integrates photometric and geometric representations from multimodal data (RGB, RGBD, and point clouds), jointly benefiting from the high-frequency texture details encoded by RGB images and metric-accurate topologies encoded by point clouds and depth maps.
Our framework employs dedicated modality-specific encoders coupled with a shared decoder architecture to map multi-modality information into a unified triplane-structured latent representation.
To effectively harness the complementary strengths of multi-modal inputs, we introduce the reconstruction-based mechanism in the diffusion model that explicitly guides the model to discern and optimally utilize the distinctive advantages of each modality.  To avoid distortion at the large elevation angle and raise the robustness and performance of multi-modal coding, we adopt auxiliary 2D (\textit{e.g.}, RGB, mask, depth loss) and 3D supervision (SDF loss). This collaborative multi-modal coding further facilitates efficient generative modeling which trains a lightweight latent diffusion model conditioned on input images. The well-structured latent representation space established by our multimodal framework enables streamlined generative model training, requiring neither intricate hyperparameter adjustments nor extensive optimization cycles. Extensive evaluations on standard benchmarks~\cite{deitke2023objaverse, downs2022google} demonstrate that, by leveraging multi-modality information, \methodname\ achieves competitive performance with recent state-of-the-art methods in generating high-quality 3D assets from images, even when trained on small-scale datasets illustrate in Fig.~\ref{fig:teaser}. Beyond exploring the potential of current 3D data, our multimodal encoding architecture inherently supports the tokenization of real-world multi-modal inputs through its extensible design. This framework provides a novel and promising pathway to systematically address the challenge of 3D training data scarcity.

Our contributions could be summarized as:
\begin{itemize}
	\item We propose \methodname, a Collaborative Multi-Modal Coding framework that fuses geometric and photometric information from multiple modalities into a unified latent space for high-quality 3D generation, using modality-specific encoders and a shared decoder.
	
	\item We introduce a hybrid 2D image-space and 3D geometric space loss for fast training and robust learning of multi-modal coding.
	
	\item Building on this multi-modal coding, we develop a generative framework that exploits the complementary strengths of different modalities while mitigating modality-specific weaknesses and the ambiguity through a tailored reconstruction loss for 3D generation.
	
	\item Extensive evaluations on well-known datasets~\cite{downs2022google, deitke2023objaverse, wu2023omniobject3d} demonstrate the superiority of our method in image-to-3D tasks quantitatively and qualitatively.
\end{itemize}

\begin{figure*}[t]
	\centering

	\includegraphics[width=1\textwidth]{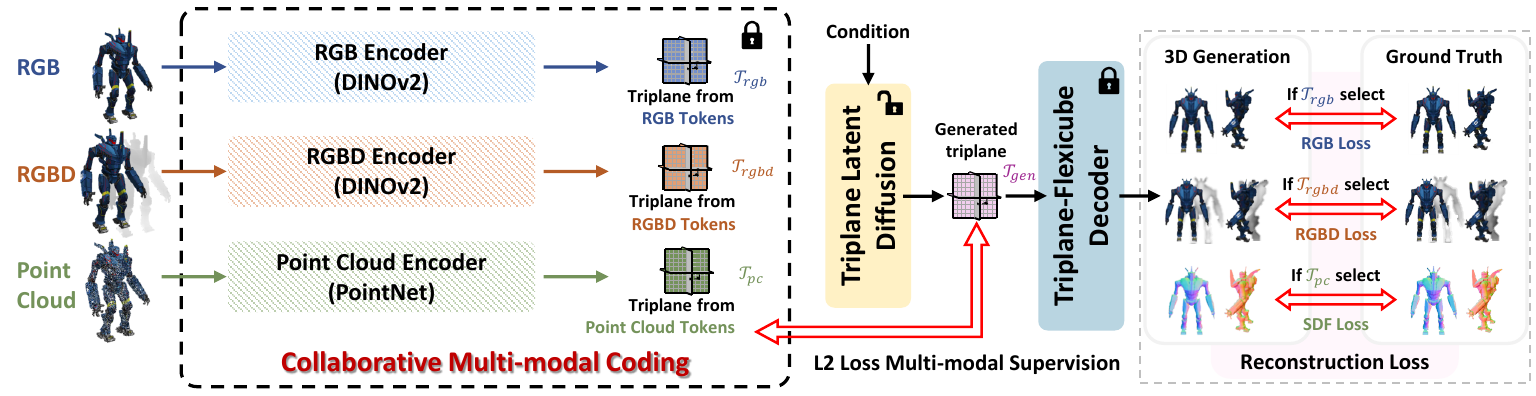}
	
	\caption{\textbf{Overview of our \methodname.} To extract the unique attributes of multi-modal triplanes and avoid their specific weakness, we introduce the loss\_2, \textit{i.e.}, reconstruction loss during training. It can guide our generative model to leverage the strength of multi-modalities coding, thereby achieving promising performance in 3D modeling. }
	
	\label{fig:pipeline}
	
\end{figure*}

\section{Related Work}

\subsection{Optimization-based methods}

As one of the most representative works in 3D generation, DreamFusion~\cite{poole2022dreamfusion} introduces SDS loss into optimization. It achieves impressive performance in 3D generation by utilizing pre-trained 2D diffusion priors. However, despite various improvements~\cite{lin2023magic3d,li2024focaldreamer,wang2024prolificdreamer, tang2023dreamgaussian}, SDS-based methods continue to face the multi-face Janus problem and low optimization efficiency, particularly when creating a large number of 3D assets.

\subsection{Feed-forward methods}
\subsubsection{3D-aware GANs} 
Before the breakthrough of the diffusion model, GAN-based methods achieve competitive performance in 3D generation. As the pioneering approach, 3D-GAN~\cite{wu2016learning} tries to learn latent representation converted via voxel. To improve the performance, EG3D~\cite{eg3d} proposes a popular representation, Triplane, which keeps a satisfactory balance between performance and efficiency. Furthermore, GET3D~\cite{get3d} adopts a two-branch framework to enrich the details in geometry.

\subsubsection{3D-native diffusion models} As one of the powerful generative models, diffusion model has demonstrated its impressive potential in the 3D generation task. Point-E~\cite{nichol2022point} proposes a diffusion model based on point cloud which is flexible but lacks the capability of representing watertight and solid surface of 3D content. To handle this, MeshDiffusion~\cite{liu2023meshdiffusion} and DiffRF~\cite{muller2023diffrf} adopt mesh and voxel as the representation. Although explicit representation has superiority in querying and evaluating, those explicit 3D representation methods are memory-expensive when using high-resolution grids. Different from explicit representation, the neural field-based methods~\cite{nam20223d, jun2023shap} adopt an implicit way to encode the geometry and texture information. Despite low memory loading in high-resolution grids, it will cost more time to evaluate. Compared with pure explicit or implicit methods, the hybrid representation triplane achieves a promising trade-off between efficiency and performance~\cite{wang2023rodin,shue20233d,cao2023large,cao2024difftf++}. Recently, TRELLIS~\cite{xiang2024structured} proposes a powerful 3D generation framework that uses large-scale high-quality 3D data and 2D\&3D reference input for reconstruction. Based on it, PhysXGen~\cite{cao2025physx} and PhysX-Anything~\cite{cao2025physxanything} introduces the first physical-grounded and simulation-ready 3D generative method. Despite introducing multi-modal data, the early fuse framework limits its scalability across different modalities and still suffers from the scarcity of 3D data. To avoid this problem, we integrate the strength of multi-modal data in the later module. In this way, our model not only utilizes the potential of current 3D data to raise the performance but also inherently supports the tokenization of heterogeneous multi-modal inputs through its extensible design.

\subsubsection{Reconstruction-based methods} 
Recently, LRM-based methods~\cite{openlrm, wei2024meshlrm, liu2024meshformer, zhang2024geolrm, xu2024grm} have gained significant attention for their impressive performance and efficiency. Based on a pure transformer-based network, LRM~\cite{lrm} can reconstruct the triplane from a single input image. Building on LRM, Instant3D~\cite{li2023instant3d} extends LRM to the multi-view model. By inputting the multi-view images generated from the multi-view diffusion model~\cite{shi2023mvdream, wang2023imagedream}, Instant3D achieves impressive performance from a single image. To raise the performance further, LGM \cite{tang2025lgm} replaces triplane by Gaussian splatting representation. Despite superior texture performance, Gaussians are limited in explicit geometry modeling and high-quality surface extraction. To deal with this, InstantMesh~\cite{xu2024instantmesh} adopts a mesh-based representation, \textit{i.e.}, Flexicube~\cite{shen2023flexible}. However, most of those methods are based on a single modality data which cannot provide sufficient information for high-quality 3D generation. To boost the performance effectively, we opt to introduce more comprehensive reference information. Therefore, we develop a multi-modal encoding framework to capture and leverage the unique advantages of different modalities for enhancing 3D generative performance. Besides direct performance improvement, our model can also be applied to 3D data extension that converts other modality data into 3D triplanes to enrich the details and diversity of generated 3D assets.

\begin{figure*}[t]
		\centering

        \includegraphics[width=1\linewidth]{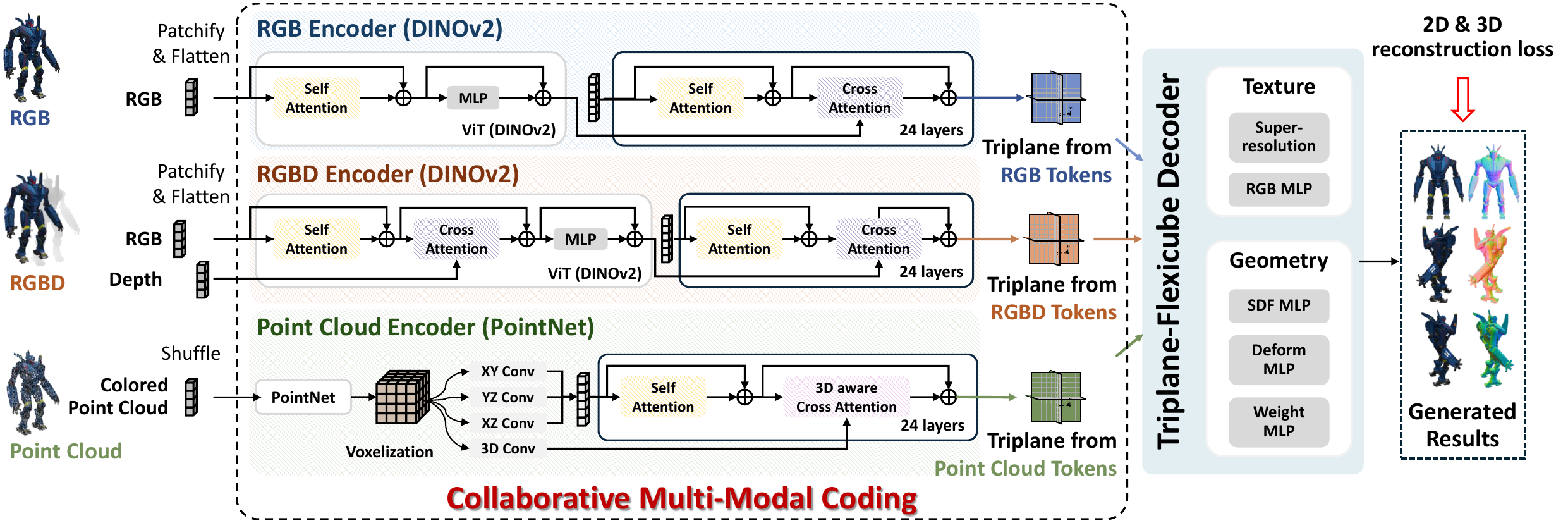}

		\caption{\textbf{Detailed structure of our Collaborative Multi-Modal Coding.} The proposed Collaborative Multi-Modal Coding can be tokenized from each of the modalities (\textit{i.e.} RGB, RGBD, Point Clouds) using different encoders, shown as the three branches above. By adopting a share-weight triplane-flexicube decoder, the coding (\textit{i.e.} corresponding triplanes) from different modalities collaboratively share a joint latent space. }
		
		\label{fig:details}
		
	\end{figure*}
\section{Methods}

The overview of our \methodname~ is shown in Fig.~\ref{fig:pipeline}. 
Our method starts from \textbf{1) Collaborative Multi-Modal Coding} which integrates multi-modal data into a unified space, followed by \textbf{2) triplane latent diffusion model} that performs generative modeling on top of the superiority of multi-modal data and embeddings.

\subsection{Collaborative multi-modal coding}
As the core component of \methodname, Collaborative Multi-Modal Coding consists of (1) modality-specific encoders and (2) a shared decoder. Specifically, the encoder has three branches, \textit{i.e.}, RGB, RGB-D, and point cloud. These encoders exploit the dense texture information in RGB and the spatial details in depth and point clouds, while the shared decoder ensures that multi-modal inputs are projected into a unified latent space.

Moreover, since our Collaborative Multi-Modal Coding is trained on public 3D datasets, it can be used to alleviate the scarcity of 3D data for generative tasks. By converting heterogeneous RGB images, RGB-D images, and point cloud datasets into a common triplane-based latent space, we can introduce diverse training data and enhance the capacity to model complex 3D structures.

For clarity, we denote three different encoders as $\mathcal{E}_\mathrm{rgb}$, $\mathcal{E}_\mathrm{rgbd}$, and $\mathcal{E}_\mathrm{pointcloud}$, respectively. The detailed structures of encoders are illustrated in Fig~\ref{fig:details}. Inspired by~\cite{triposr}, we adopt a pure transformer architecture to encode the RGB image into triplane. As for the RGBD images, we split and patchify the RGB image and depth image using different convolution layers. To maintain the robustness and performance of the encoder, we utilize and fine-tune the pre-train foundation model, \textit{i.e.}, DINOv2. By introducing the depth image via the cross-attention module and residual connection, we can obtain an RGBD tokenizer efficiently. Therefore, the triplane from the RGB and RGBD images can be formulated as:
\begin{align}
		&\mathcal{T}_\mathrm{rgb}=\mathcal{E}_\mathrm{rgb}(\mathcal{R}_\mathrm{rgb}),  \\
		&\mathcal{T}_\mathrm{rgbd}=\mathcal{E}_\mathrm{rgbd}(\mathcal{I}_\mathrm{rgb},\mathcal{R}_{d}),
\end{align}
where $\mathcal{R}_\mathrm{rgb}\in \mathbb{R}^{3\times 512 \times 512}$ and $\mathcal{R}_{d}\in \mathbb{R}^{1\times 512 \times 512}$ represent the inputted RGB and depth images.

To explicitly represent the spatial information of point clouds, we opt to transform the point cloud features into voxel-like representations. Let us denote the input point cloud as $\mathcal{P}\in \mathbb{R}^{n \times 6}$ where the first three channels represent the normalized XYZ coordinates of points while the last three channels represent the RGB value of corresponding points. After shuffling the point clouds, we adopt the most classic point cloud feature backbone, \textit{i.e.}, PointNet~\cite{NIPS2017pointnet}, to extract point cloud features as follows:
\begin{equation}
	\begin{aligned}
		&\mathcal{\hat{P}}=\mathcal{E}_{pn}(\mathcal{P}), \;\; \mathcal{\hat{P}}\in \mathbb{R}^{n \times 64}\,,
	\end{aligned}
\end{equation}
where $\mathcal{E}_{pn}$ represent the PointNet. 
To establish the relationship between point cloud features and the voxel grid, we partition the unit cube into $N\times N\times N$ grids (denoted as $\mathcal{C}$). The features in each voxel grid can be queried from corresponding 3D cubes. Therefore, the voxel-like feature can be calculated by:
\begin{equation}
	\begin{aligned}
		&\mathcal{F}_{pc}[:,i,j,k]=\dfrac{1}{L}\sum^{N}_{l=0} \large{(}\,\mathcal{\hat{P}}[l,\,:]\;\; \Large{\cap} \;\; \mathcal{C}[i,\,j,\,k]\,\large{)}.
	\end{aligned}
\end{equation}
Then, we adopt three different convolution layers to project the voxel-like feature $\mathcal{F}_{\mathrm{pc}}[i,j,k]$ to three planes by:
\begin{equation}
	\hat{\mathcal{T}}_{pc}=\mathrm{Cat}
	{\Large{\{}}
	f_\mathrm{Conv}^{yz}(\overline{\mathcal{F}^X_{pc}}), \, f_\mathrm{Conv}^{xz}(\overline{\mathcal{F}^Y_{pc}}), \, f_\mathrm{Conv}^{xy}(\overline{\mathcal{F}^Z_{pc}})
	{\Large{\}}},
\end{equation}
where $\overline{\mathcal{F}^k_{pc}}, k=X, Y, Z$ and $f_\mathrm{Conv}$ represent calculating mean value on the k axis and convolution layer, respectively. Besides, to effectively maintain the spatial information of the colored point clouds, we introduce a 3D convolution layer and a 3D-aware cross-attention layer in the transformer encoder. Thus, the triplanes from point clouds can be formulated as:
\begin{equation}
	\begin{aligned}
		\mathcal{T}_{pc}=\mathcal{E}_\mathrm{tf}(\hat{\mathcal{T}}_{pc},\;\mathcal{F}_{pc})
	\end{aligned}
	~ ,
\end{equation}
where $\mathcal{E}_\mathrm{tf}$ denotes the transformer model for point clouds.

To project multi-modal triplanes to the same latent space, we adopt a share-weight decoder. To accelerate the training process, we first train the RGB branch including the RGB encoder and shared decoder. Then, we initialize other encoders with RGB triplanes and fix the parameters of the decoder to train the RGBD encoder and point cloud encoder. 

\begin{figure}[t]
	\centering

	\includegraphics[width=1\linewidth]{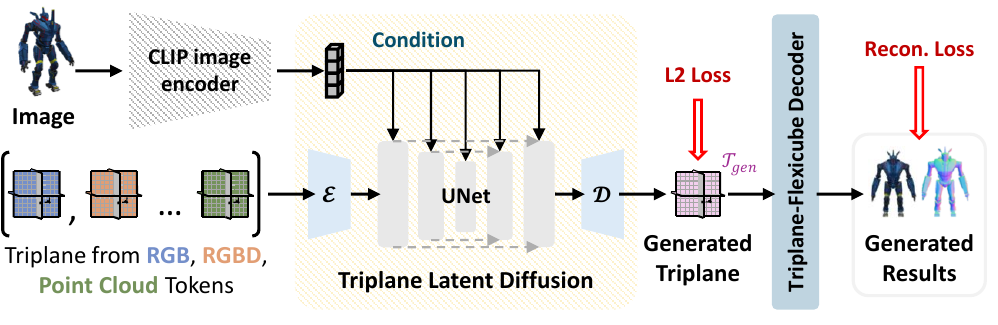}
	\vspace{-3mm}
	\caption{\textbf{Training pipeline of our triplane latent diffusion model.} It can harness and integrate the distinctive attributes of various modalities via reconstruction loss, thereby producing 3D assets enriched with rich texture and finely detailed structures.}
	\label{fig:detail_diff}
	
\end{figure}

To improve the performance in terms of geometry and texture, we introduce a hybrid 2D image-space loss and 3D geometric loss. Similar to prior work, we use reconstruction loss in the 2D image space of rendering, depth, and mask. 
Besides, we introduce a 3D loss based on the Signed Distance Function (SDF) to optimize the geometry of 3D assets directly.
To accelerate the convergence and avoid the negative influence caused by the unbalanced positive-negative ratio, we calculate the BCEloss on positive and negative areas, respectively. 
Please note that to avoid unnecessary triangle and high-frequency noise, the regularization function is also included in our loss function following previous work~\cite{shen2023flexible}. Thus, the loss function can be formulated as:
\begin{equation}
	\begin{aligned}
		\mathcal{L}_{\mathrm{code}}=\mathcal{L}_{\mathrm{rgb}}(\hat{I}_i,I_i^{gt})+\lambda_\mathrm{d}\mathcal{L}_{\mathrm{d}}&(\hat{I}_d,I_d^{gt})+\lambda_\mathrm{reg}\mathcal{L}_\mathrm{reg}\\
		+\lambda_\mathrm{mask}\sum_{i}\Vert\hat{M}_i-M_i^{gt}\Vert_2^2&+\lambda_\mathrm{sdf}\mathcal{L}_{\mathrm{sdf}}(\hat{S},S^{gt}),\\
	\end{aligned}
\end{equation}
where $I_i$, $I_i^{gt}$, $I_D$, $I_D^{gt}$, $\hat{M}_i$, and $M_i^{gt}$ represent the rendered RGB image, ground-truth image, rendered depth image, ground-truth depth image, rendered mask, ground-truth mask of $i$-th view. 
\begin{equation}
	\begin{aligned}
		\mathcal{L}_{\mathrm{sdf}}(\hat{S},S^{gt})=(\mathcal{L}_\mathrm{BCE}(\hat{S}_{p},S^{gt}_{p})+\mathcal{L}_\mathrm{BCE}(\hat{S}_{n},S^{gt}_{n})),
	\end{aligned}
\end{equation}
where $\mathcal{L}_\mathrm{BCE}$ is the BCE loss, $\hat{S}_{p}$, $S^{gt}_{p}$, $\hat{S}_{n}$, and $S^{gt}_{n}$ denote the positive SDF area, ground-truth positive SDF area, negative SDF area, and ground-truth negative area.

\subsection{Triplane latent diffusion model}

After obtaining the multi-modal triplanes from collaborative Multi-Modal Coding, we adopt a typical diffusion model to learn the superiority of different modality data. Note that to reduce the difficulty of training the diffusion model, we introduce a variational autoencoder (VAE) model for spatial compression of multi-modal triplanes. Besides, during the optimization of VAE, we use the KL loss to constrain the distribution of the compressed triplanes. Compared to directly integrating the VAE model with our collaborative Multi-Modal Coding, using two separate models allows for more parameters and larger batch sizes during training while also reducing the convergence difficulty of our Multi-Modal Coding. Using the RGB triplane as an example, the VAE loss function is defined as follows:
\begin{equation}
	\begin{aligned}
		\mathcal{L}_{\mathrm{vae}}=\mathcal{L}_\mathrm{MSE}(\hat{\mathcal{T}}_\mathrm{rgb}^{c},\mathcal{T}_\mathrm{rgb}^{c}))+\lambda_{kl}\mathrm{D}_{KL}(N(\mu,\sigma)|N(0,1))
	\end{aligned}
	~ ,
\end{equation}
where $\mathcal{L}_\mathrm{MSE}$ is the MSE loss, $N(\mu,\sigma)$ is the distribution of the reconstructed compressed triplanes ($\hat{\mathcal{T}_{i}^{c}}$) and $\mathcal{T}_{i}^{c}$ represent the compressed triplanes.

After obtaining the VAE model, we then train the diffusion model. The pipeline of our diffusion model is shown in Fig.~\ref{fig:detail_diff}. Considering the wide application of image-to-3D models, we use image embedding extracted by the CLIP model as the condition to optimize the diffusion model. A standard U-Net diffusion model with ResBlocks and downsampling and upsampling layers is utilized. We train the triplane diffusion model to predict the noise $\epsilon$ added to the compressed triplane features, applying an L2 loss to the prediction. Besides, to extract and leverage the attributes of different modalities, we add the reconstruction loss for corresponding modalities as follows:
\begin{equation}
	\begin{aligned}
		\mathcal{L}_{\mathrm{diff}}&=\mathbb{E}_{t\sim [1,T],~\epsilon\sim N(0,1)}[\Vert f_{\theta}(z^t,t,c)-\epsilon\Vert]+\lambda_{\mathrm{rec}}\mathcal{L}_{\mathrm{rec}}\\
	\end{aligned}
	~ ,
\end{equation}
where T, $z^t$, and c are the time steps of the diffusion model, noised triplane features in $t$-th step, and conditional image embedding while $f_{\theta}$ is the triplane diffusion model. $ \mathcal{L}_{\mathrm{rec}}$ is corresponding to the modality of triplanes as follows:
\begin{align}
	\mathcal{L}_{\mathrm{rec}}=
	\left\{
	\begin{aligned}
		&\mathcal{L}_{\mathrm{rgb}}(\mathcal{I}^\mathrm{\prime},\mathcal{I}^{gt}), \,~~~~~~~~~~~~~~~~~~~~~~~~~\text{if}~~\mathcal{T}~\text{is}~\mathcal{T}_\mathrm{rgb}
		\\
		&\mathcal{L}_{\mathrm{rgb}}(\mathcal{I}^\mathrm{\prime},\mathcal{I}^{gt})+\mathcal{L}_{\mathrm{d}}(\mathcal{I}^\mathrm{\prime}_{d},\mathcal{I}^{gt}_{d}), ~~\text{if}~~\mathcal{T}~\text{is}~\mathcal{T}_\mathrm{rgbd}
		\\
		& \mathcal{L}_{\mathrm{sdf}}(\hat{S}^\mathrm{\prime},S^{gt}), ~~~~~~~~~~~~~~~~~~~~~~~~~~\text{if}~~\mathcal{T}~\text{is}~\mathcal{T}_\mathrm{pc}
	\end{aligned}
	\right.
\end{align}
where $\mathcal{I}^\mathrm{\prime}$, $\mathcal{I}^\mathrm{\prime}_{d}$, and $\hat{S}^\mathrm{\prime}$ are RGB image, depth image, and SDF value rendered from generated triplane $\mathcal{T}_\mathrm{gen}$.

\begin{figure}[t]
	\centering

	\includegraphics[width=0.5\textwidth]{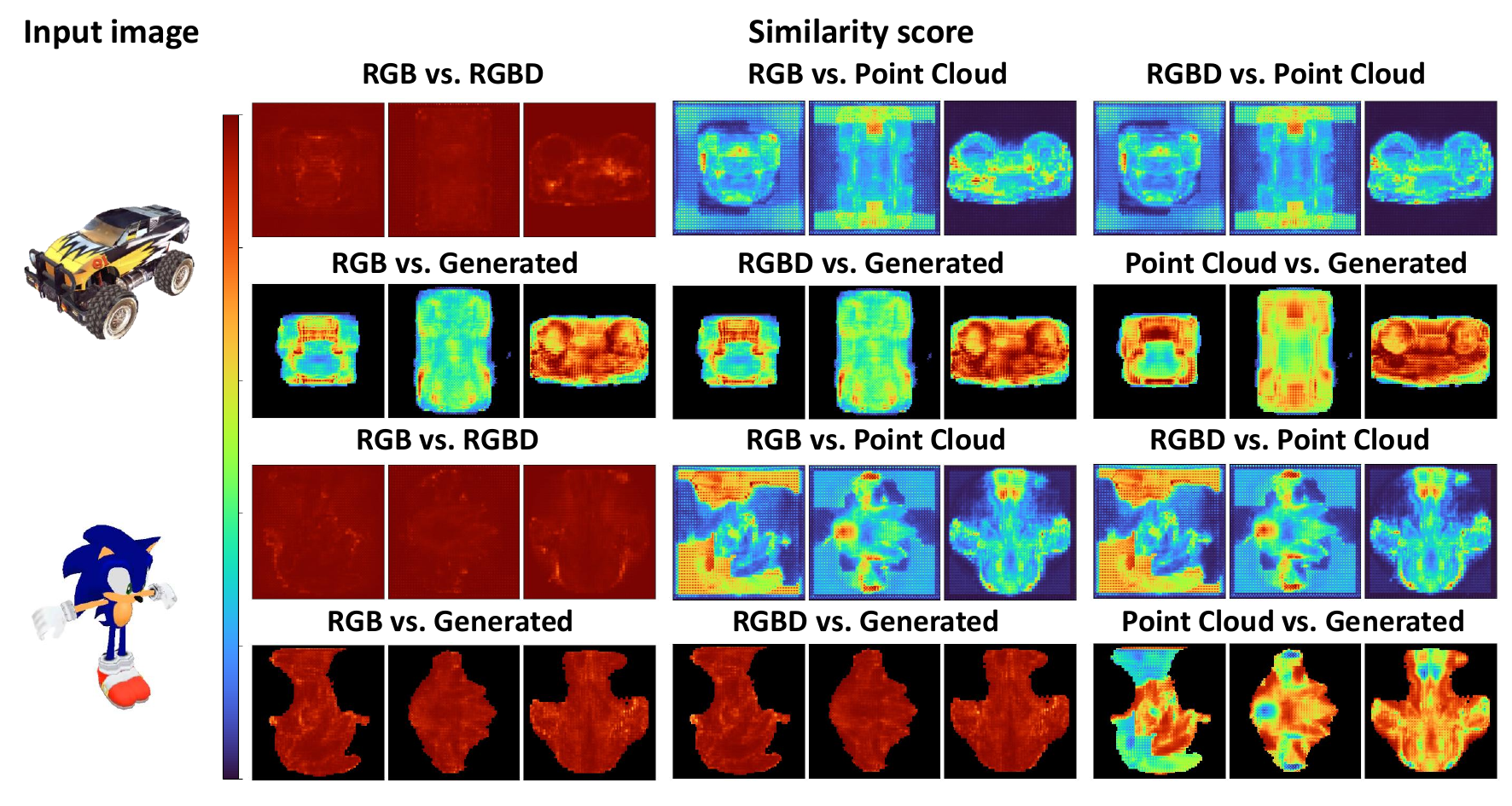}
	\vspace{-6.5mm}
	\caption{Similarity scores between the multi-modality triplanes and the generated triplane. The results indicate that the original multi-modality triplanes exhibit limited interdependence, whereas our generated triplane effectively integrates and leverages information from all modalities.}\label{fig:comp}
	
	\vspace{-3.5mm}
	
\end{figure}

To validate the modality interaction in our method, we visualize and compare the similarity score of multi-modality triplanes and our generated results in Fig.~\ref{fig:comp}. The results show that the original multi-modality triplanes (top) lack explicit correlation. By integrating these triplanes via diffusion, our generated triplane exhibits clear correlations with all modalities (bottom).

\begin{figure}[t]
	\centering
	\includegraphics[width=0.5\textwidth]{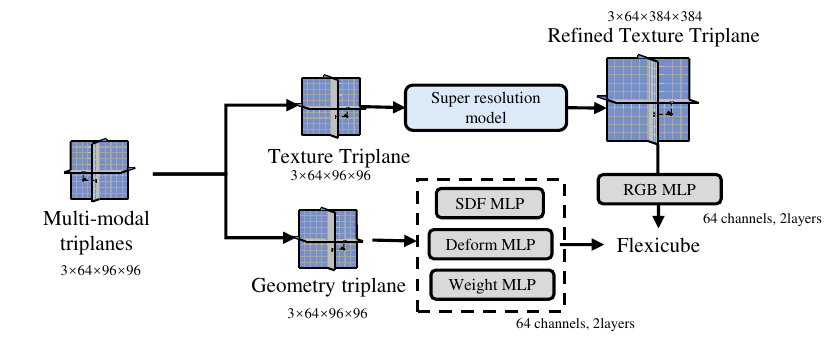}
	\vspace{-6.5mm}
	\caption{\textbf{Details of the decoder in our collaborative multi-modal coding.} Leveraging the lightweight decoder, our model efficiently and effectively transforms the triplane into a colored mesh.}
	\vspace{-2.5mm}
	\label{fig:dec}
\end{figure}

\begin{figure*}[t]
	\centering	
	
	\includegraphics[width=1\linewidth]{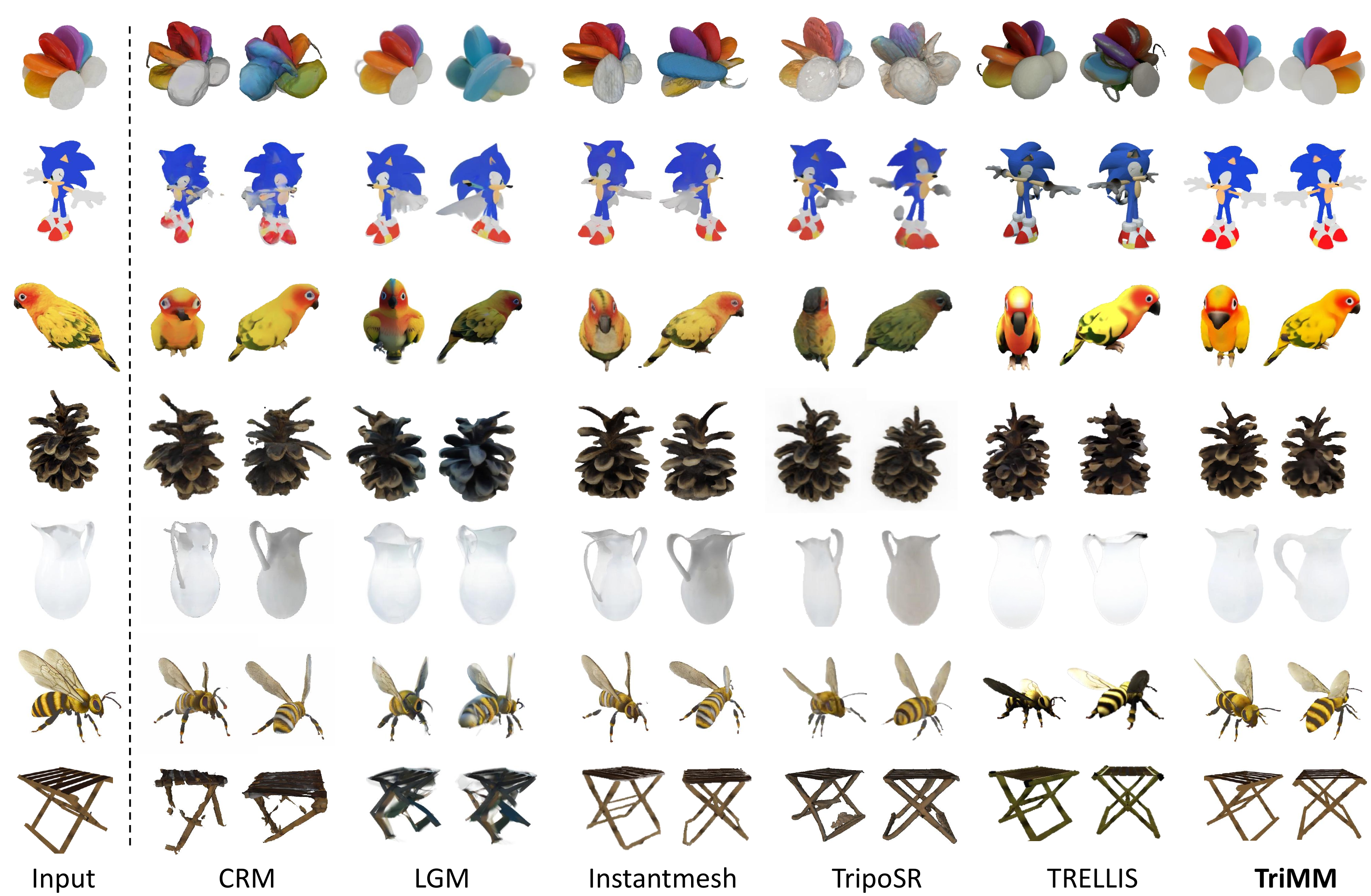}
	\vspace{-2.5mm}
	\caption{\textbf{Qualitative results.} We compared our \methodname\ with other methods on image-to-3D. Thanks to our multi-modal coding the collaboratively marries photometric and geometric information in the unified triplane space, our \methodname\ achieves impressive generative performance, especially for fine-grained geometric details like wings and hairs.}
	
	\label{fig:quali}
\end{figure*}

\definecolor{1s}{RGB}{255,179,179} 
\definecolor{2s}{RGB}{255,204,204} 
\definecolor{3s}{RGB}{255,230,230} 
\section{Experiments}
\begin{figure*}[t]		
	\includegraphics[width=1\linewidth]{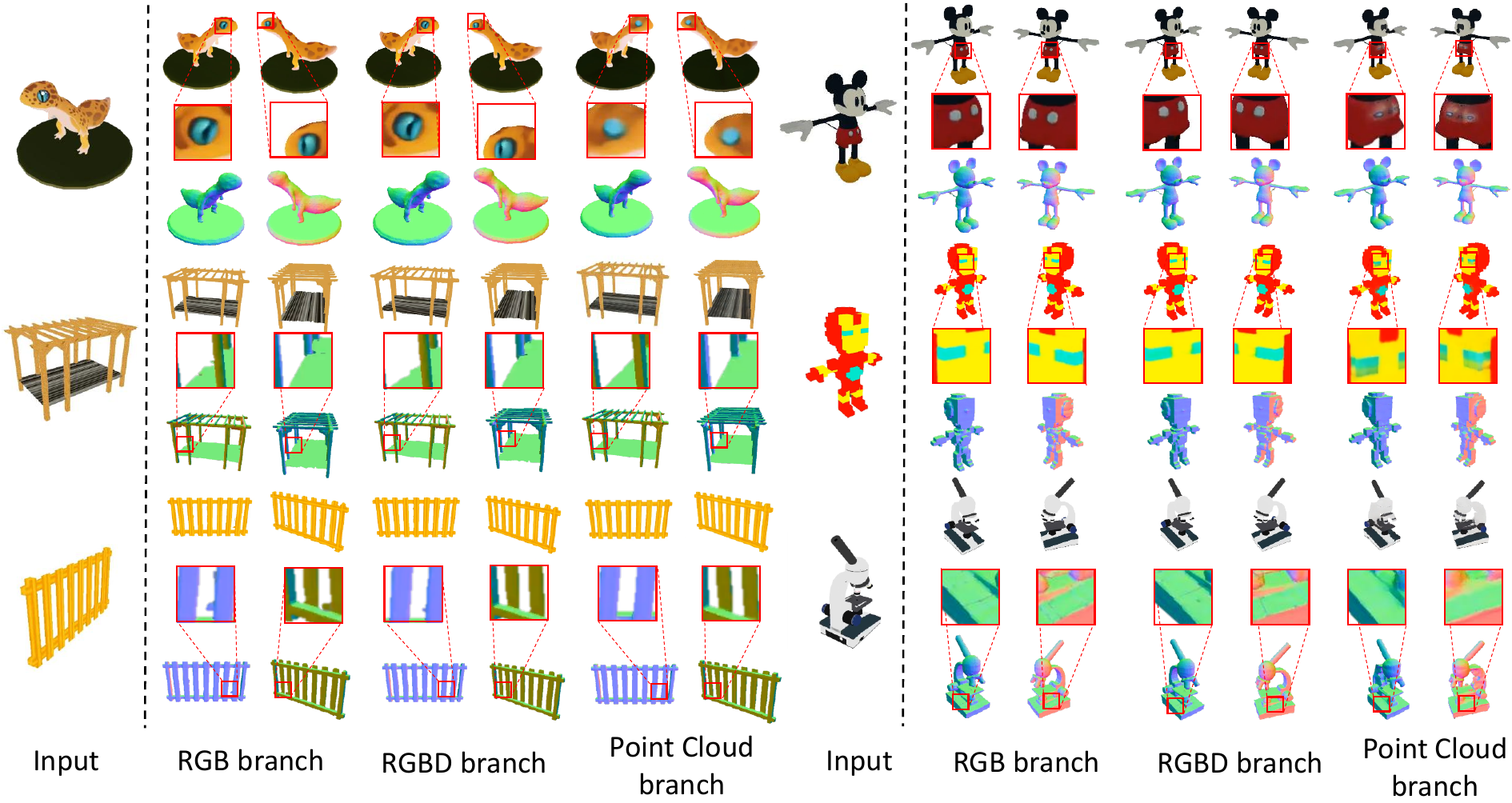}
	\vspace{-3.5mm}
	\caption{\textbf{Qualitative comparison among branches based on different modalities.} It shows the strengths of different modality data for 3D modeling. The point cloud branch excels at capturing geometry, whereas the RGB and RGBD branches perform better at modeling texture.}
	
	\label{fig:modalitycom}
\end{figure*}

\begin{figure}[t]
	\centering	
	
	\includegraphics[width=0.5\textwidth]{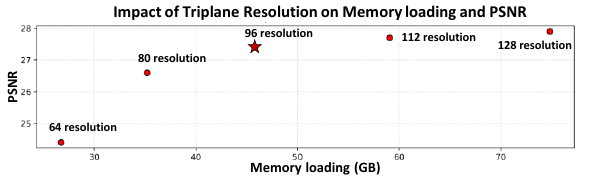}
	\caption{Study of triplanes with different resolutions. To achieve a satisfactory trade-off between efficiency and performance, we adopt the resolution of 96 for our generated results.}\label{fig:star}

\end{figure} 

\begin{figure}[t]
	\centering

	\includegraphics[width=0.5\textwidth]{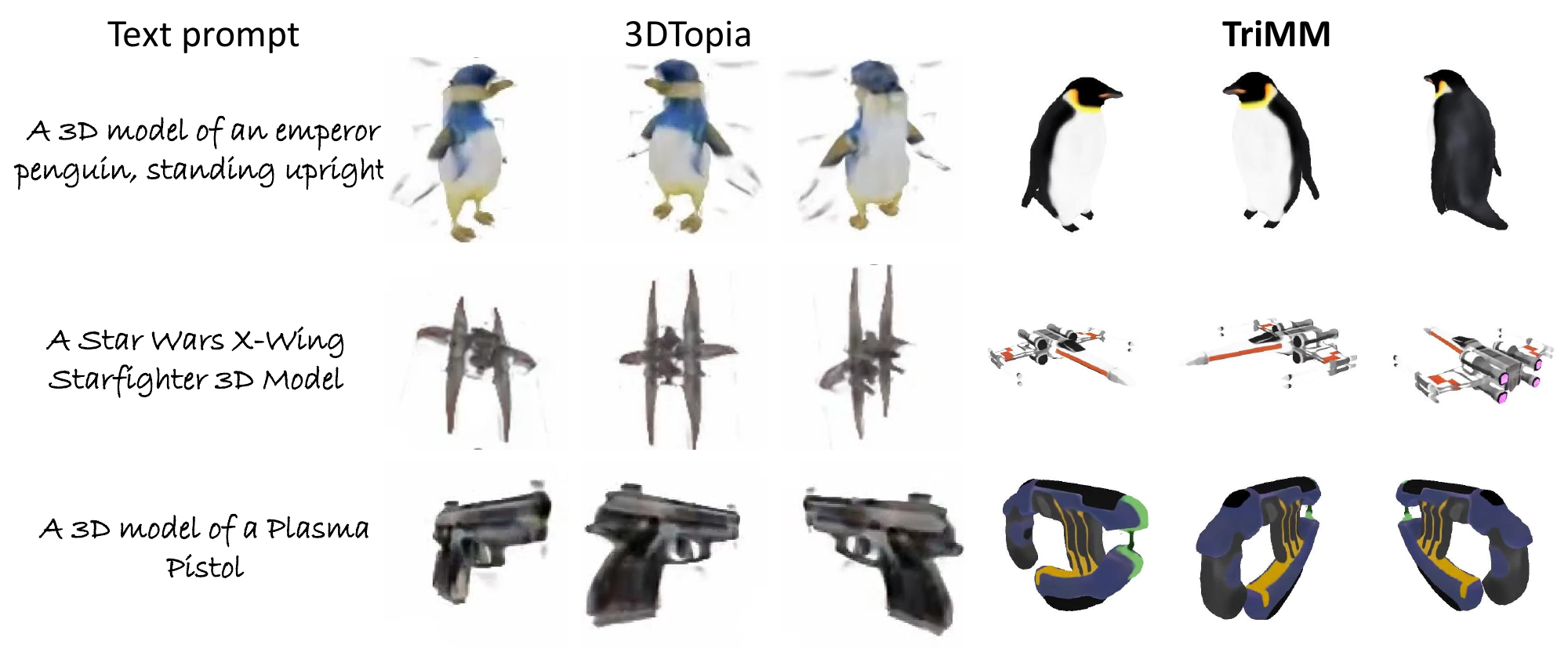}
	\vspace{-10pt}
	\caption{Qualitative comparison with text-conditional generation method.}\label{fig:eval}

\end{figure}
\subsection{Implementation details}
\noindent\textbf{Training data.} Our proposed collaborative multi-modal coding is trained on a high-quality subset of Objaverse~\cite{deitke2023objaverse} including around 80K 3D objects. We adopt the same rules as~\cite{tang2025lgm} to filter the low-quality 3D objects. Then, we render the RGB, mask, and depth images from 8 random views at the resolution of 512 $\times$ 512 for training. As for the input reference image, we adopt the rendered images of 512 $\times$ 512 in elevation angles between [-15,30] degrees. In the reconstruction evaluation, we randomly sample 2000 objects from Objaverse. 

\noindent\textbf{Evaluation data.} To evaluate the generalization of our model in unseen objects, we conduct qualitative and quantitative experiments on Google Scanned Objects~\cite{downs2022google} (GSO) and OmniObject3D~\cite{wu2023omniobject3d}. The GSO dataset includes about 1K 3D objects while OmniObject3D~\cite{wu2023omniobject3d} involves more than 6K diverse 3D objects. We used 400 objects in the GSO dataset and 2000 3D objects in OmniObject3D during the evaluation. Besides, we sample 1000 3D objects from Objaverse for reconstruction evaluations. More implementation details including the structure, training details, and evaluation metrics are released in the supplementary.

\noindent\textbf{Evaluation metrics.}
In our experiments, we evaluate the texture and geometry of generated 3D assets. For texture evaluation, we normalize the mesh and sample 16 random views from a unit sphere to calculate the mean Peak Signal-to-Noise Ratio (PSNR). Besides, we also introduce the CLIP score to measure the consistency between the generated results and the input prompts. To evaluate the quality of geometry, we calculate the standard shape metrics of Chamfer Distance (CD) and F-score (FS) with thresholds of 0.05 and 0.1.

\noindent\textbf{Details of \methodname}
The decoder consists of a super-resolution module and 4 MLP networks shown in Fig.~\ref{fig:dec}. We adopt a traditional convolution network~\cite{lim2017enhanced} to enrich the texture of the generated objects. Besides, to maintain the robustness of unified triplane space, we only use two layers to project the triplane feature to SDF, deform, weight, and RGB. 

\begin{table}[t]
	\centering
	
	\renewcommand\tabcolsep{3pt}
	
	\caption{Comparison of several modules parameters, training, and inference time.}
	\begin{tabular}{lcccc}
		\toprule
		Model name & \#params. & Training time & Inference time   \\
		\midrule
		RGB branch coding & 676.6M  & 90 hours & 3.3s \\
		RGBD branch coding & 761.9M  & \textless5 hours & 4.1s \\
		Point cloud branch coding & 631.5M & 12 hours & 4.5s \\
		VAE & 585.6M  & 24 hours & 0.8s \\
		Diffusion & 985.9M  & 24 hours & 1.2s \\

		\bottomrule
	\end{tabular}
	
	\label{tab:ab3}%
\end{table}%

\noindent\textbf{Training stages}
In this subsection, we report more details about our \methodname. We compare the parameters, training, and inference time of different modules in Table~\ref{tab:ab3}. We use 24 NVIDIA A100 GPUs to train our model. Note that the inference time is tested on NVIDIA A100 GPUs. We split our model into two stages: 1) multi-modal coding training including RGB, RGBD, point cloud tokenizer, and shared decoder; 2) Triplane latent diffusion training including VAE and image-conditional diffusion model. For the 1st stage, we first train a RGB reconstruction model similar to other LRM-based methods, \textit{e.g.}, InstantMesh~\cite{instantmesh}. Besides, to accelerate the convergence of our multi-modal coding, we use the pre-trained RGB triplane to initialize other triplanes. After obtaining all multi-modality triplanes, we train a typical latent diffusion model (2nd stage). Considering the application, we adopt a single image as the condition to train the generative model. We have conducted additional experiments with different triplane resolutions and observed that higher resolutions consistently improve reconstruction performance shown in Fig.~\ref{fig:star}. However, they also introduce substantially heavier memory consumption and greater difficulty for diffusion training. Therefore, we adopt a resolution of 96×96 in our final model as a balanced trade-off between accuracy and efficiency.

\noindent\textbf{Inference stages}
After training, we can use the diffusion model, decoder of VAE, and shared decoder of multi-modal coding to generate and render the results. Based on an inputted single RGB image, it can generate a 3D object within 4 seconds.

\subsection{Representation learning results}
In this subsection, we compare our collaborative multi-modal coding results with the reconstruction results of other LRM-based baseline methods, \textit{e.g.}, TripoSR~\cite{triposr}, LGM~\cite{tang2025lgm}, and InstantMesh~\cite{instantmesh}. Table~\ref{tab:reccom} shows the impressive reconstruction performance of our proposed modules. The comparison among the last three lines validates the claims that different modalities have their unique superiority in 3D modeling. Relying on additional depth information, the RGBD branch achieves better geometry metrics results than the RGB branch, thereby having the ability to modeling small structures which brings the gain in texture. Besides, the point cloud branch shows the best performance in geometry learning.

\begin{table}[t]
	\setlength{\tabcolsep}{8pt}
	\centering
	\caption{\textbf{Quantative results on the Objaverse~\cite{deitke2023objaverse}.} We compare the reconstruction results with other methods. Our method outperforms both photometrically and geometrically.}
	\small{
		
		\begin{tabular}{l|ccc}
			\toprule
			Methods & PSNR$\uparrow$&CD$\downarrow$&FS\texttt{@}0.05$\uparrow$  \\
			\midrule
			TripoSR~\cite{triposr}&24.67&0.0147&0.972\\
			LGM~\cite{tang2025lgm}&26.23&0.0287&0.951\\
			InstantMesh~\cite{instantmesh}&26.22&0.0124&0.987\\
			\midrule
			\textbf{RGB (Ours)}&\underline{27.81}&0.0084&\textbf{0.999}\\
			\textbf{RGBD (Ours)}&\textbf{28.32}&\underline{0.0041}&\textbf{0.999}\\
			\textbf{PointCloud (Ours)}&26.10&\textbf{0.0026}&\textbf{0.999}\\
			

			\bottomrule
		\end{tabular}
	}

	\label{tab:reccom}%
	
\end{table}%

\subsection{Triplane diffusion results}
\subsubsection{Qualitative evaluations}

\begin{figure}[t]
	\centering
	\includegraphics[width=0.5\textwidth]{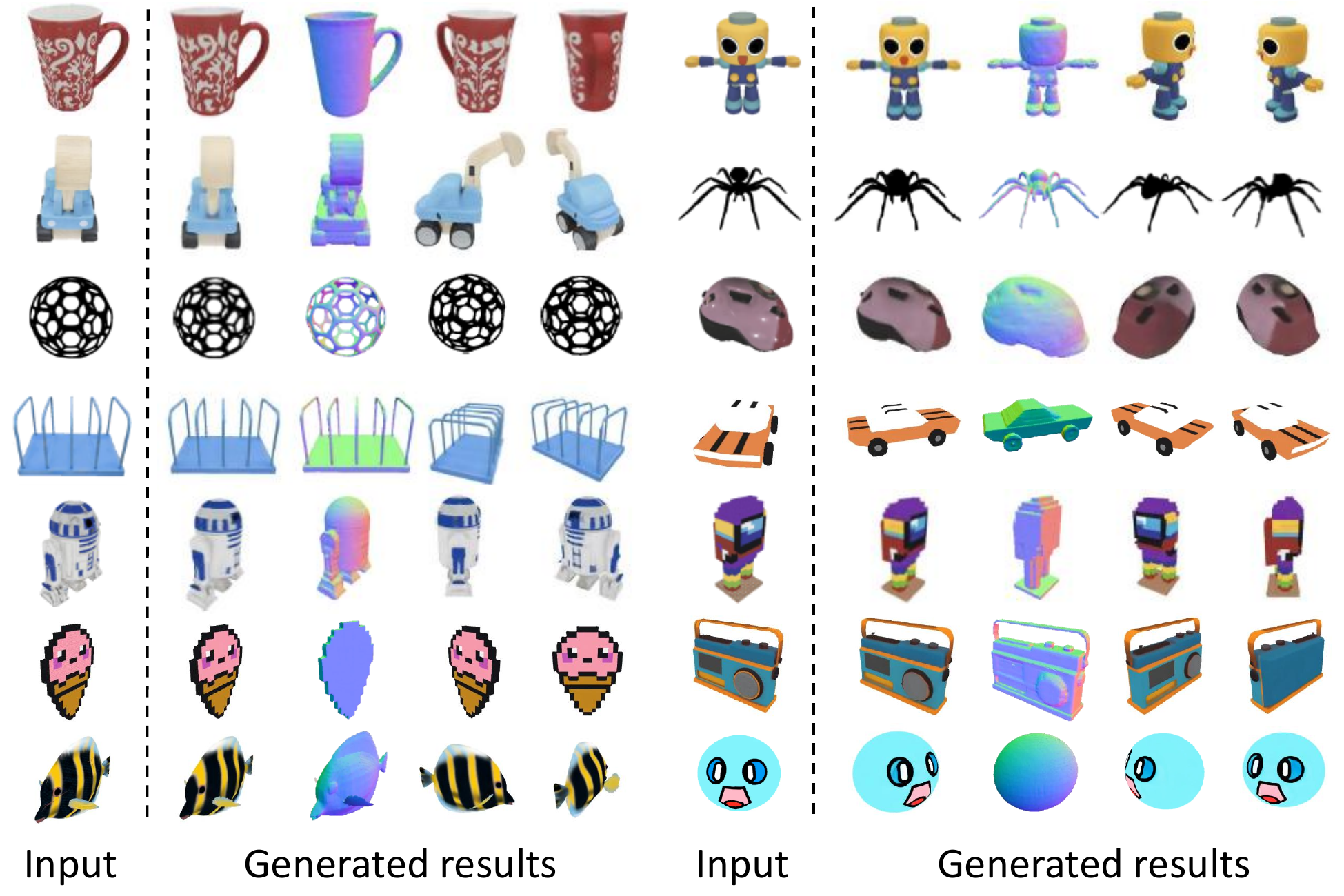}
	\caption{\textbf{More qualitative results of our proposed method.} It shows that our method can generate high-quality 3D assets with photo-realistic textures and detailed geometry}
	\label{fig:our1}
\end{figure}

In this subsection, we compared our method with our baseline and the most recently existing works including TripoSR~\cite{triposr}, CRM~\cite{wang2025crm}, LGM~\cite{tang2025lgm}, InstantMesh~\cite{instantmesh}, and TRELLIS~\cite{xiang2024structured}. Baseline represents the generative model trained only on RGB triplanes. We adopt the pre-trained model LGM (trained on 80k data), CRM (trained on 376k data), InstantMesh (trained on 270k data), and TRELLIS (trained on 500k data). We compared our \methodname\ with other SOTA methods qualitatively in Fig.~\ref{fig:quali}. We can notice that the generated 3D meshes of our \methodname\ are more stable when facing objects with different appearances and structures. TripoSR can generate a promising texture for 3D assets, especially in the frontal view. However, since the triplane representation has limited capacity to model complex geometric structures, it tends to produce overly flattened 3D shapes. Besides, since LGM, CRM, and InstantMesh are built based on a multi-view diffusion model, the performance of those methods is constrained by the performance of the diffusion model which decreases the final performance of image-to-3D. Relying on high-quality large-scale training data, TRELLIS achieves superior robustness and performance in various 3D assets. Different from integrating 2D and 3D data directly, we introduce the multi-modal coding module based on multi-modal data, thereby raising the robustness when facing complex structures and maintaining the capability of extension. Despite using smaller data, by exploring the potential of current 3D assets, our model achieves a competitive performance compared with TRELLIS. More qualitative results are shown in Fig.~\ref{fig:our1}. Although the main focus of this paper is on the image-to-3D generation task, we additionally extend our experiments to the text-to-3D setting. We use Qwen-Image~\cite{wu2025qwen} to generate conditioning images from text prompts and then apply our TriMM framework to produce 3D assets. The comparison with the text-to-3D method 3DTopia~\cite{3dtopia} demonstrates that our approach achieves impressive generative performance.

\subsubsection{Quantitative evaluations}
In this subsection, we report the quantitative results for image-to-3D on unseen samples from public datasets~\cite{downs2022google, wu2023omniobject3d}. 
As shown in Table~\ref{tab:gso} and Table~\ref{tab:omni}, \methodname\ achieves competitive quality both geometrically and photometrically. 
It is worth noting that by utilizing and leveraging the multi-modal data, our \methodname\ can achieve competitive performance of other methods trained on large-scale datasets.
We attribute this performance boost to our collaborative multimodal coding which explicitly incorporates both 2D and 3D information.


	

\begin{table}[t]
	\centering

	\renewcommand\tabcolsep{2pt}
	\caption{\textbf{Quantative results on Google Scanned Objects (GSO)~\cite{downs2022google}}. We compare \methodname\ trained on 80k data with four state-of-the-art approaches: LGM (80k), CRM (376k), InstantMesh (270k), and TRELLIS (500k) on the GSO dataset. Our method outperforms both photometrically and geometrically.}
	\small{
		\begin{tabular}{l|cc|ccc}
			
			\toprule
			\multirow{2}*{Methods}&\multicolumn{2}{c|}{Appearance} & \multicolumn{3}{c}{Geometry}\\
			& CLIP$\uparrow$&PSNR$\uparrow$&CD$\downarrow$&FS\texttt{@}0.05$\uparrow$&FS\texttt{@}0.1$\uparrow$   \\
			\midrule
			Baseline&48.2&13.37& 0.061& 0.427& 0.591\\
			
			TripoSR~\cite{triposr}&47.8&13.05& 0.096& 0.409& 0.564\\
			CRM~\cite{wang2025crm}&49.7&13.83&0.043&0.562&0.756\\
			LGM~\cite{tang2025lgm}&50.3&14.32&0.072&0.327&0.525\\
			InstantMesh~\cite{instantmesh}&50.5&14.17&0.058&0.457&0.654\\
			Trellis~\cite{xiang2024structured}&\underline{51.4}&\underline{14.27}&\underline{0.035}&\textbf{0.612}&\underline{0.775}\\
			\midrule
			\textbf{\methodname\ (Ours)}& \textbf{52.5}&\textbf{14.34}&\textbf{0.034}&\underline{0.607}&\textbf{0.786}\\
			

			\bottomrule
		\end{tabular}
	}
	
	\label{tab:gso}%
	
\end{table}%

\begin{table}[t]
	\centering
	\caption{\textbf{Quantative results on OmniObject3D~\cite{wu2023omniobject3d}}. We compare \methodname\ trained on 80k objects with four state-of-the-art approaches: LGM (80k), CRM (376k), InstantMesh (270k), and TRELLIS (500k) on 1700 unseen samples from the OmniObject3D dataset. Our method outperforms both photometrically and geometrically.}
	
	\renewcommand\tabcolsep{2pt}
	\small{
		\begin{tabular}{l|cc|ccc}
			\toprule
			\multirow{2}*{Methods}&\multicolumn{2}{c|}{Appearance} & \multicolumn{3}{c}{Geometry}\\
			& CLIP$\uparrow$&PSNR$\uparrow$&CD$\downarrow$&FS\texttt{@}0.05$\uparrow$&FS\texttt{@}0.1$\uparrow$   \\
			\midrule
			Baseline& 65.3 & 12.14&0.131& 0.259& 0.427\\
			TripoSR~\cite{triposr}& 66.4 & 12.64&0.129& 0.262& 0.439\\
			CRM~\cite{wang2025crm}&{67.2}&13.22&0.109&0.283&0.477\\
			LGM~\cite{tang2025lgm}&66.3&13.29&0.131&0.253&0.421\\
			InstantMesh~\cite{instantmesh}&66.1&13.37&0.105&0.331&0.512\\
			Trellis~\cite{xiang2024structured}&\underline{67.3}&\underline{13.50}&\underline{0.102}&\underline{0.374}&\underline{0.556}\\
			\midrule
			\textbf{\methodname\ (Ours)}& \textbf{67.4}&\textbf{14.13}&\textbf{0.096}&\textbf{0.379}&\textbf{0.561}\\
			

			\bottomrule
		\end{tabular}
	}

	\label{tab:omni}%
\end{table}%


	
	
	
\definecolor{1s}{RGB}{255,179,179} 
\definecolor{2s}{RGB}{255,204,204} 
\definecolor{3s}{RGB}{255,230,230} 
\begin{table}[t]
	\setlength{\tabcolsep}{3.2pt}
	\centering
	\caption{\textbf{Ablation studies about multi-modal coding.} By introducing specific multi-modal data, our \methodname\ can improve the performance in aspects corresponding to the attributes of each modality. Top-3 are highlighted in different colors.}

	\small{
		\begin{tabular}{cccc|cc|cc}
			\toprule
			\multirow{2}*{\footnotesize{RGB}}&\multirow{2}*{\footnotesize{RGBD}}& \footnotesize{Point} &{\footnotesize{Recons.}}&\multicolumn{2}{c|}{\footnotesize{Appearance}} & \multicolumn{2}{c}{\footnotesize{Geometry}}\\
			&& \footnotesize{Cloud} & \footnotesize{Loss} &\footnotesize{CLIP$\uparrow$} &\footnotesize{PSNR$\uparrow$} &\footnotesize{CD$\downarrow$} &\footnotesize{FS\texttt{@}0.05$\uparrow$}  \\
			\midrule
			\Checkmark&\XSolidBrush&\XSolidBrush&\XSolidBrush& 55.2&13.6&0.116&0.362\\
			\Checkmark&\XSolidBrush&\XSolidBrush&\Checkmark& 58.3&14.5&0.084&0.425\\
			\XSolidBrush&\Checkmark&\XSolidBrush&\Checkmark& 58.4&14.5&0.046&0.512\\
			\XSolidBrush&\XSolidBrush&\Checkmark&\Checkmark&56.1&13.6&0.023&0.613\\
			\midrule
			\Checkmark&\Checkmark&\XSolidBrush&\Checkmark&\underline{60.1}&\underline{15.4}&0.047&0.511\\
			\Checkmark&\XSolidBrush&\Checkmark&\Checkmark&59.3&14.6&0.021&0.622\\
			\XSolidBrush&\Checkmark&\Checkmark&\Checkmark&58.1&14.2&\textbf{0.015}&\underline{0.639}\\
			\midrule
			\Checkmark&\Checkmark&\Checkmark&\XSolidBrush&57.5&14.3&0.048&0.503\\ \Checkmark&\Checkmark&\Checkmark&\Checkmark&\textbf{64.8}&\textbf{16.6}&\textbf{0.015}&\textbf{0.641}\\ 

			\bottomrule
		\end{tabular}
	}

	\label{tab:ab}%
\end{table}%

\begin{figure}[t]
	\centering

	\includegraphics[width=1\linewidth]{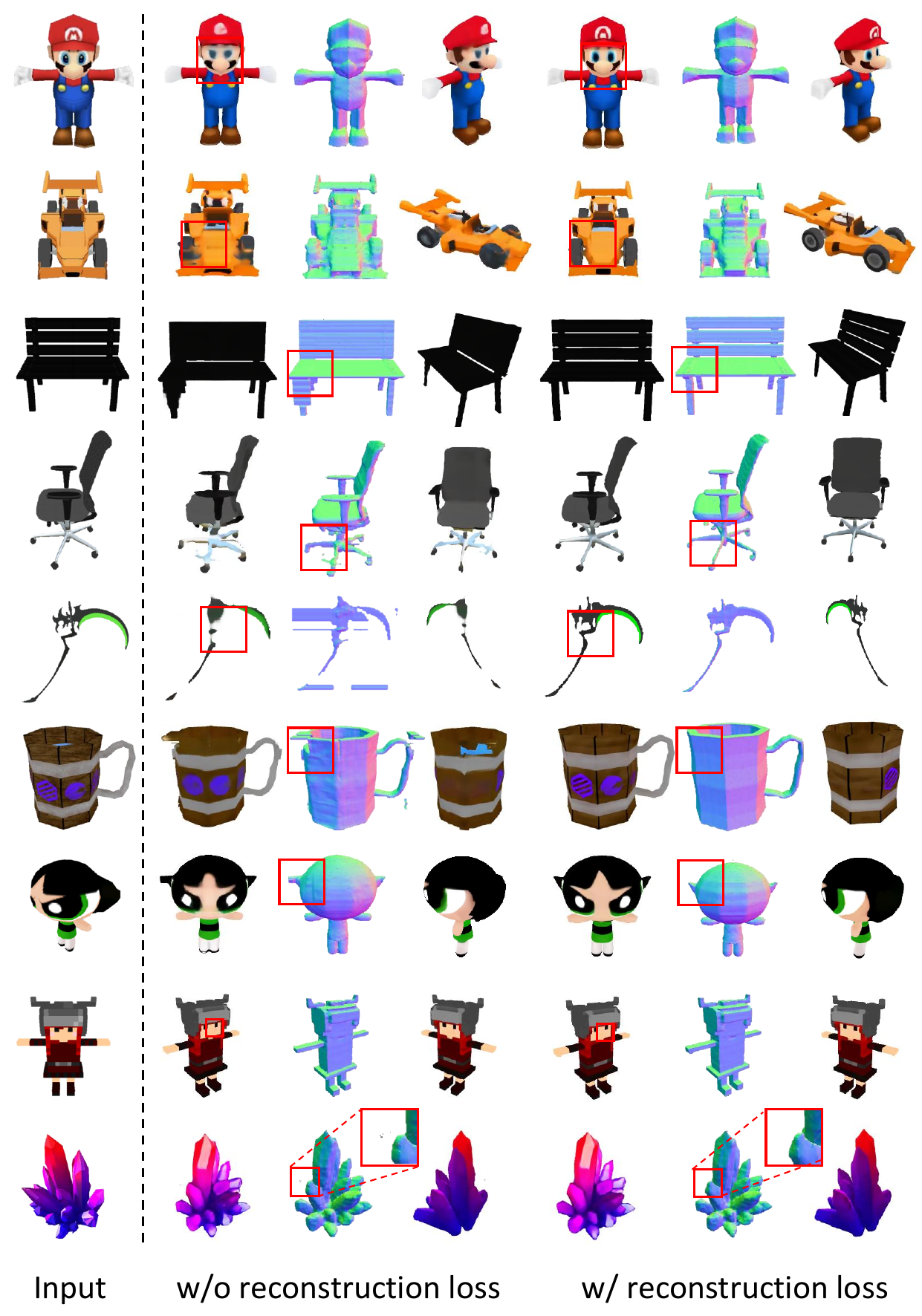}
	\vspace{-5mm}
	\caption{\textbf{Ablation studies of reconstruction loss.} By introducing the reconstruction loss, our model can avoid the weakness of specific modality, thereby enhancing the generative performance further.}
	
	\label{fig:loss}
	
	\vspace{-10pt}
\end{figure}	

\subsection{Ablation studies}
In this subsection, we discuss the effectiveness of introducing multi-modal data, reconstruction loss, 2D\&3D supervision, and VAE model. 

\subsubsection{Analysis about multi-modal data.} As shown in the first three lines of Table.~\ref{tab:ab}, we can notice that developing generative models based on the RGB, RGBD, or point cloud can lead to different performances of texture and geometry. Integrating two different modalities can mitigate the limitations of generative models based on a single modality. For instance, training the generative model based on RGB and colored point cloud can bring improvement in geometry compared to the model based on RGB and enrich the texture compared to the model based on colored point cloud. 
Since the multi-modality triplanes are mixed during training, the distribution of different modality triplanes will affect the final generative performance. Therefore, the model using RGBD and point cloud sources introduces more geometric information (depth and 3D coordinates). It causes the model to pay more attention to geometry, thereby impeding the improvement in texture quality. Finally, by leveraging the different attributes of all three multi-modal data, \textit{i.e.}, the dense texture information of RGB image, depth information of RGBD image, and abundant geometric information of colored point clouds, our \methodname\ achieves impressive overall performance. 

\subsubsection{Analysis about reconstruction loss.}
We also conduct experiments on reconstruction loss in our model and we adopt the RGB, RGBD, and point clouds as our multimodal sources. To effectively harness the strength of multimodal data for 3D modeling, we try to integrate multimodal triplanes in the generative model by the reconstruction loss. It aims to optimize the generative model based on different superiority of different modalities. As shown in Table.~\ref{tab:ab}, the model with reconstruction loss achieves a significant improvement. Also, we compare the model without reconstruction loss qualitatively in Fig.~\ref{fig:loss}. The model without reconstruction loss can merely 

\begin{figure}[t]
	\centering
	\includegraphics[width=0.5\textwidth]{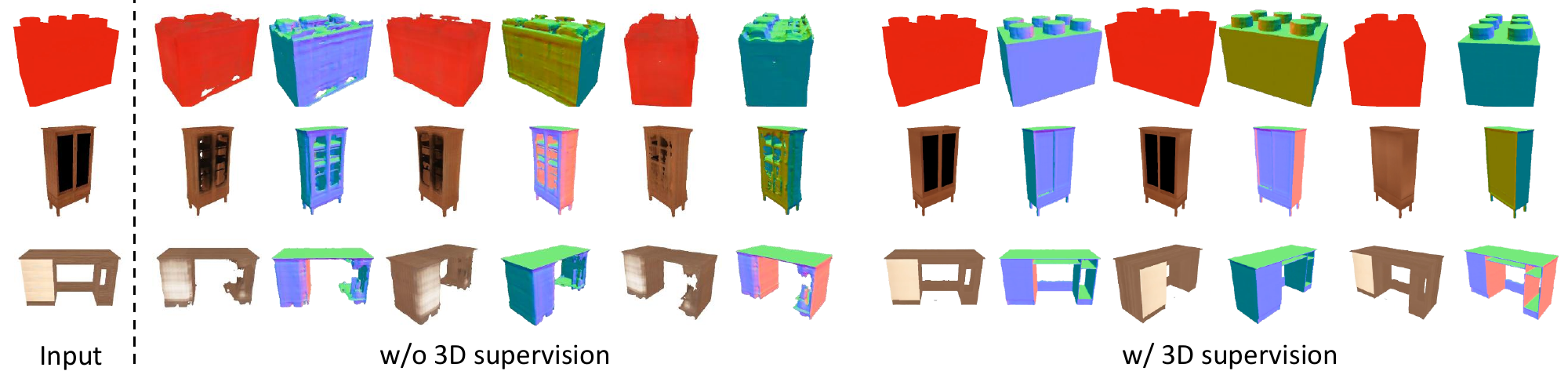}
	\vspace{-3mm}
	\caption{\textbf{Comparison between model using 2D supervision and model using 2D\&3D supervision.} By leveraging 3D supervision, our method can directly learn geometric structures and effectively avoid artifacts.}
	\label{fig:ab3}
	
	\includegraphics[width=0.5\textwidth]{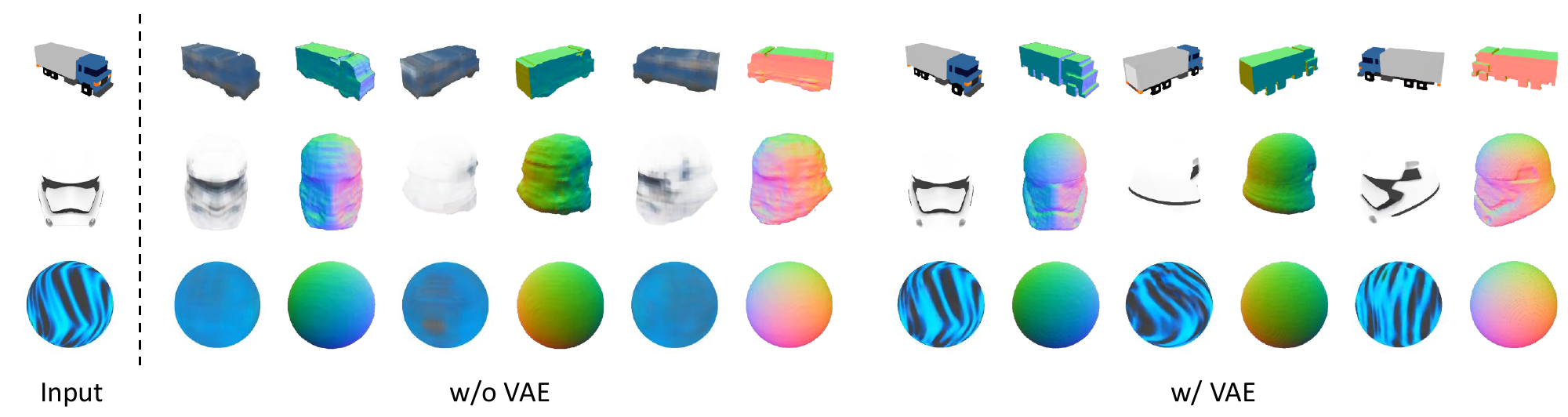}
	
	\caption{\textbf{Ablation studies on the VAE in Triplane latent diffusion model.} We analyze the effectiveness of VAE. It shows that introducing VAE to compress the multi-modal triplanes can accelerate the convergence of the generative model.}
	
	\label{fig:abl2}
	
	\includegraphics[width=0.5\textwidth]{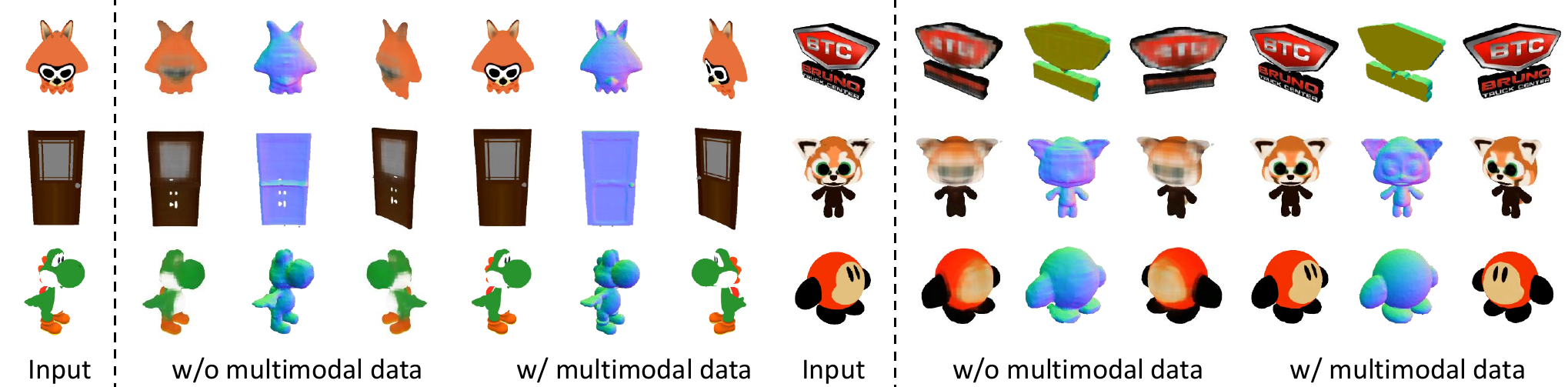}
	\caption{Comparison between models with and without utilizing multimodal data. By Introducing multimodal data, the generative model is more stable in various conditions. }
	\label{fig:other}
	\vspace{-10pt}
\end{figure}

\subsubsection{Analysis of 2D/3D supervision}
In this section, we validate the effectiveness of incorporating 2D/3D hybrid supervision. As shown in Fig.~\ref{fig:ab3}, introducing 3D supervision enables our model to better capture geometric details, particularly when handling 3D content with complex structures. Furthermore, it reduces training time and accelerates convergence. The quantitative comparison, presented in Table~\ref{tab:ab4}, shows that the model trained with both 2D and 3D supervision significantly outperforms the model using only 2D supervision in terms of texture and geometry quality. These qualitative and quantitative results strongly demonstrate the effectiveness of 3D supervision.
\begin{table}[b]
	\centering
	
	\renewcommand\tabcolsep{3pt}
	\small
	\caption{\textbf{Ablation studies about 2D/3D supervision in reconstruction model on the subset of Objaverse.} By adopting 3D supervision, our collaborative multi-modal coding gains a promising improvement.}
	\begin{tabular}{lccc}
		\toprule
		Methods & PSNR$\uparrow$&CD$\downarrow$&FS\texttt{@}0.05$\uparrow$ \\
		\midrule
		
		w/o 3D suerpvision&24.26&0.015&0.961\\
		
		\midrule
		\textbf{w/ 3D supervision (ours)} &28.12&0.0048&0.999\\

		\bottomrule
	\end{tabular}
	
	\label{tab:ab4}%
\end{table}%

\subsubsection{Discussion on VAE.} We conduct experiments on the introduction of the VAE model. By introducing the VAE model, our \methodname\ can be built on the compressed latent features encoded by multi-modal triplanes. As shown in Fig.~\ref{fig:abl2}, we compared the diffusion results of the same epoch. It proves that introducing a VAE model can accelerate the convergence speed of our \methodname\. Besides, the quantitative results in Table~\ref{tab:vae} validate the effectiveness of the VAE model.

		
		

	

\begin{table}[b]
	\centering
	
	\renewcommand\tabcolsep{2pt}
	\caption{\textbf{Ablation studies about VAE}. We compare the generative results with and without the VAE model. By introducing the VAE module, the learned features become more compact and informative, thereby leading to improved performance.}
	\begin{tabular}{l|cc|ccc|}
		
		\toprule
		\multirow{2}*{Methods}&\multicolumn{2}{c|}{Appearance} & \multicolumn{3}{c|}{Geometry}\\
		& CLIP$\uparrow$&PSNR$\uparrow$&CD$\downarrow$&FS\texttt{@}0.05$\uparrow$&FS\texttt{@}0.1$\uparrow$   \\
		\midrule
		w/o VAE&{43.2}&{12.71}&{0.108}&{0.387}&{0.521}\\
		\midrule
		\textbf{w/ VAE (ours)}& 52.1&{14.83}&{0.034}&{0.612}&{0.761}\\
		

		\bottomrule
	\end{tabular}
	
	\label{tab:vae}%
\end{table}%

\subsection{Further analysis}
\subsubsection{Multi-modal data to 3D.}
To validate our statement about the multi-modality, we visualize the reconstruction results of RGB-to-3D, RGBD-to-3D, and point cloud-to-3D in Fig~\ref{fig:modalitycom}. By introducing more geometric information, the point cloud branch can avoid the artifacts effectively. However, since it uses sparse RGB information, it is more likely to ignore the details of the texture while the RGB and RGBD branch can handle this problem by introducing dense RGB information.

\subsubsection{Introducing multi-modalities for 3D generation}

In this paper, we propose a novel collaborative multimodal coding module to project multimodal data into triplane latent space. It can not only extract and utilize the attributes of multimodal data but also provide a new direction for extending trainable 3D data. To validate the feasibility of this application, we conduct the experiments on a subset of Objaverse~\cite{deitke2023objaverse}, WildRGB-D~\cite{xia2024rgbd}. Note that since Objaverse focussed more on isolated objects, we utilize the RGBD and point cloud data of isolated objects rather than scenes. In this experiment, we adopt a two-step training strategy: 1) adopting triplanes from all multimodal data (Objaverse\& WildRGB-D) to pre-train the generative model; and 2) adopting high-quality triplanes from Objaverse to enhance the performance of the generative model. In this way, we can improve the quality and diversity of generated 3D assets. As shown in Fig.~\ref{fig:other}, the model introducing multimodal sources is more stable compared to the one without using multimodal sources when facing diverse 3D objects. Besides, we compare the generative performance of w/o and w/ additional datasets shown in Table~\ref{tab:otherdata}. It proves that introducing RGBD data can improve the quality and diversity of the generative model. Therefore, we think that by introducing existing large-scale multi-modalities data and high-quality 3D data, 3D generation capabilities can be effectively enhanced. We will continue on this problem in our future work.

\subsubsection{User studies.}
\begin{figure}[t]
	\centering	
	
	\includegraphics[width=1\linewidth]{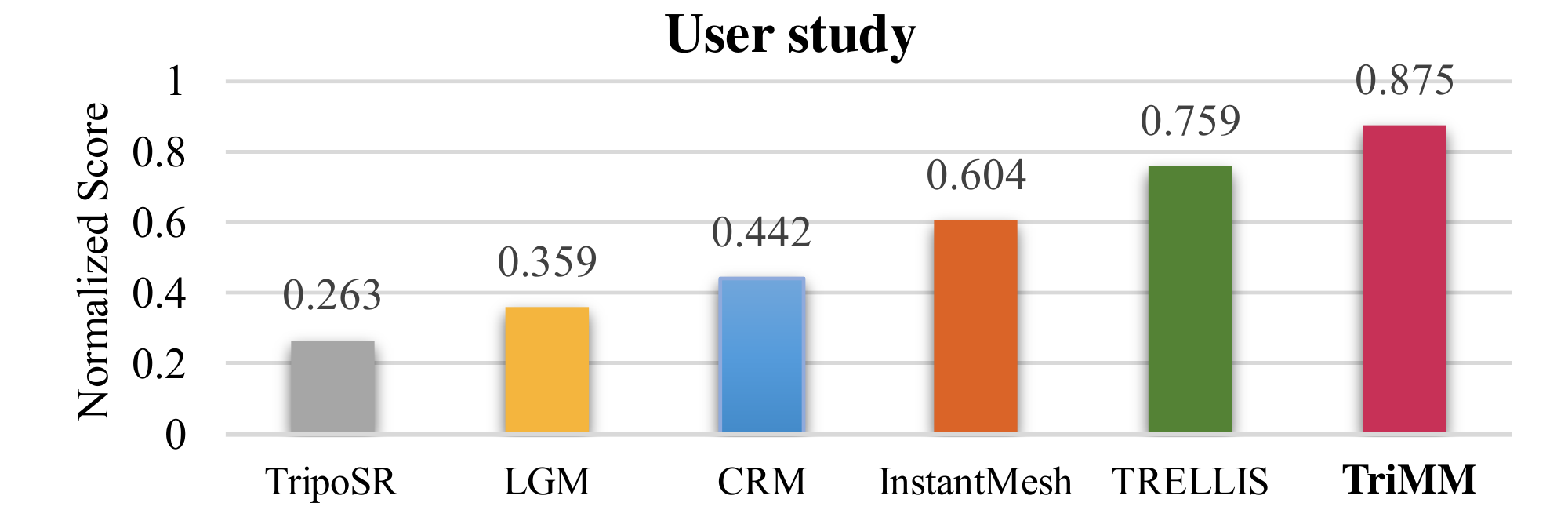}
	\vspace{-2mm}
	\caption{\textbf{User study results on the overall quality of 3D generation from different methods.} The normalized score proves the promising performance of our method.}
	
	\label{fig:user}
	
\end{figure}

To evaluate our methods more comprehensively, we also conduct user studies on the quality of generated 3D assets. Specifically, we render 360-degree rotating videos
for five generative methods ~\cite{triposr,wang2025crm,tang2025lgm,instantmesh,xiang2024structured} and our proposed method. There are 48 videos totally in our evaluation, including car, vegetable, food, house, and so on. Each volunteer will rate the generated results of those six methods anonymously from 1 to 6. We normalize and report the score in Fig.~\ref{fig:user}. It shows that our methods are more aligned with human preferences.

\subsubsection{Limitation and future work}

Despite the promising results, our method still has several limitations and opens up multiple avenues for future work. First, the triplane representation suffers from inefficient information utilization and is inherently memory-bound when scaling to higher resolutions, which restricts its ability to capture very fine-grained geometric and appearance details. A promising direction is to explore alternative or hybrid 3D representations, such as a combination of triplanes with Gaussian splatting, neural implicit fields, or multi-resolution hierarchical variants. In particular, Gaussian splatting has shown strong advantages in terms of rendering quality and efficiency with compact point-based primitives, but it is less straightforward to encode structured articulation and physical attributes. Designing a hybrid representation that leverages the structured factorization of triplanes and the expressiveness and efficiency of Gaussian splats may offer better trade-offs between expressiveness, memory consumption, and differentiable rendering, thereby further improving reconstruction quality. Second, our current use of multi-modality supervision is constrained by the configuration of existing datasets. Many point cloud datasets provide only geometric information, while most RGB-D datasets are scene-centric rather than object-centric. Although object segmentation masks help localize target objects, they cannot fundamentally compensate for the limited resolution and noise in RGB-D sensing. In future work, we plan to investigate more flexible strategies for exploiting heterogeneous and partially missing modalities, as well as to incorporate richer real-world multi-modality data, which we believe will further enhance the robustness of our model and expand the pool of usable 3D assets.

\begin{table}[t]
	\centering

	\renewcommand\tabcolsep{3pt}
	\caption{\textbf{Further analysis on additional multi-modalities data.} By introducing additional data, our method achieves a promising improvement in terms of geometry and texture.}
	{
		\begin{tabular}{lccc}
			\toprule
			Methods & CLIP$\uparrow$&CD$\downarrow$&FS\texttt{@}0.05$\uparrow$ \\
			\midrule
			
			w/o additional dataset&62.4&0.022&0.637\\
			
			\midrule
			\textbf{w/ additional dataset} &64.9&0.0018&0.645\\

			\bottomrule
		\end{tabular}
	}
	
	\label{tab:otherdata}%
\end{table}%


\section{Conclusion}\label{sec:conclusion}

Generally, RGB image contains dense texture information while point clouds and depth images include more geometrical details. Those special attributes can provide abundant reference information for robust 3D generation. However, most existing works are developed based on single-modality data, neglecting the prospective direction of improvement by introducing multi-modal data. Therefore, In this work, we propose the novel collaborative multi-modal coding for 3D generation, \textit{i.e.}, \methodname. It consists of two stages: 1) collaborative multi-modal coding and 2) triplane latent diffusion model. The former can capture the special attributes of multi-modal data and integrate them into a unified latent space. Based on the multi-modal coding module, we use a triplane latent diffusion model to generate high-quality 3D assets. By introducing a special reconstruction, it can leverage the strength of specific modalities while avoiding their weakness. Comprehensive experiment results on public datasets validate the impressive performance of our \methodname. Furthermore, \methodname\ highlights a promising direction for developing a unified multimodal 3D generation model by extending multi-source data.

	\ifCLASSOPTIONcompsoc
	\section*{Acknowledgments}
	\else
	\section*{Acknowledgment}
	\fi
	
	This study is supported by the National Key R\&D Program of China (2022ZD0160201), and Shanghai Artificial Intelligence Laboratory. This study is also supported by the Ministry of Education, Singapore, under its MOE AcRF Tier 2 (MOE-T2EP20223-0002), NTU NAP, and under the RIE2020 Industry Alignment Fund – Industry Collaboration Projects (IAF-ICP) Funding Initiative, as well as cash and in-kind contribution from the industry partner(s). $\hfill\blacksquare$
	
	\ifCLASSOPTIONcaptionsoff
	\newpage
	\fi

\bibliographystyle{IEEEtran}
\bibliography{ref}

@article{nichol2022point,
	title={Point-E: A System for Generating 3D Point Clouds from Complex Prompts},
	author={Nichol, Alex and Jun, Heewoo and Dhariwal, Prafulla and Mishkin, Pamela and Chen, Mark},
	journal={arXiv preprint arXiv:2212.08751},
	year={2022}
}

@article{wu2023omniobject3d,
	title={OmniObject3D: Large-Vocabulary 3D Object Dataset for Realistic Perception, Reconstruction and Generation},
	author={Wu, Tong and Zhang, Jiarui and Fu, Xiao and Wang, Yuxin and Ren, Jiawei and Pan, Liang and Wu, Wayne and Yang, Lei and Wang, Jiaqi and Qian, Chen and others},
	journal={arXiv preprint arXiv:2301.07525},
	year={2023}
}

@article{lrm,
	title={Lrm: Large reconstruction model for single image to 3d},
	author={Hong, Yicong and Zhang, Kai and Gu, Jiuxiang and Bi, Sai and Zhou, Yang and Liu, Difan and Liu, Feng and Sunkavalli, Kalyan and Bui, Trung and Tan, Hao},
	journal={arXiv preprint arXiv:2311.04400},
	year={2023}
}

@article{instantmesh,
	title={Instantmesh: Efficient 3d mesh generation from a single image with sparse-view large reconstruction models},
	author={Xu, Jiale and Cheng, Weihao and Gao, Yiming and Wang, Xintao and Gao, Shenghua and Shan, Ying},
	journal={arXiv preprint arXiv:2404.07191},
	year={2024}
}

@article{triposr,
	title={Triposr: Fast 3d object reconstruction from a single image},
	author={Tochilkin, Dmitry and Pankratz, David and Liu, Zexiang and Huang, Zixuan and Letts, Adam and Li, Yangguang and Liang, Ding and Laforte, Christian and Jampani, Varun and Cao, Yan-Pei},
	journal={arXiv preprint arXiv:2403.02151},
	year={2024}
}

@inproceedings{rombach2022high,
	title={High-resolution image synthesis with latent diffusion models},
	author={Rombach, Robin and Blattmann, Andreas and Lorenz, Dominik and Esser, Patrick and Ommer, Bj{\"o}rn},
	booktitle={Proceedings of the IEEE/CVF conference on computer vision and pattern recognition},
	pages={10684--10695},
	year={2022}
}

@article{blattmann2023stable,
	title={Stable video diffusion: Scaling latent video diffusion models to large datasets},
	author={Blattmann, Andreas and Dockhorn, Tim and Kulal, Sumith and Mendelevitch, Daniel and Kilian, Maciej and Lorenz, Dominik and Levi, Yam and English, Zion and Voleti, Vikram and Letts, Adam and others},
	journal={arXiv preprint arXiv:2311.15127},
	year={2023}
}

@inproceedings{deitke2023objaverse,
	title={Objaverse: A universe of annotated 3d objects},
	author={Deitke, Matt and Schwenk, Dustin and Salvador, Jordi and Weihs, Luca and Michel, Oscar and VanderBilt, Eli and Schmidt, Ludwig and Ehsani, Kiana and Kembhavi, Aniruddha and Farhadi, Ali},
	booktitle={Proceedings of the IEEE/CVF Conference on Computer Vision and Pattern Recognition},
	pages={13142--13153},
	year={2023}
}

@article{deitke2024objaverse,
	title={Objaverse-xl: A universe of 10m+ 3d objects},
	author={Deitke, Matt and Liu, Ruoshi and Wallingford, Matthew and Ngo, Huong and Michel, Oscar and Kusupati, Aditya and Fan, Alan and Laforte, Christian and Voleti, Vikram and Gadre, Samir Yitzhak and others},
	journal={Advances in Neural Information Processing Systems},
	volume={36},
	year={2024}
}

@article{schuhmann2022laion,
	title={Laion-5b: An open large-scale dataset for training next generation image-text models},
	author={Schuhmann, Christoph and Beaumont, Romain and Vencu, Richard and Gordon, Cade and Wightman, Ross and Cherti, Mehdi and Coombes, Theo and Katta, Aarush and Mullis, Clayton and Wortsman, Mitchell and others},
	journal={Advances in Neural Information Processing Systems},
	volume={35},
	pages={25278--25294},
	year={2022}
}

@article{poole2022dreamfusion,
	title={Dreamfusion: Text-to-3d using 2d diffusion},
	author={Poole, Ben and Jain, Ajay and Barron, Jonathan T and Mildenhall, Ben},
	journal={arXiv preprint arXiv:2209.14988},
	year={2022}
}

@article{3dtopia,
	title={3dtopia: Large text-to-3d generation model with hybrid diffusion priors},
	author={Hong, Fangzhou and Tang, Jiaxiang and Cao, Ziang and Shi, Min and Wu, Tong and Chen, Zhaoxi and Yang, Shuai and Wang, Tengfei and Pan, Liang and Lin, Dahua and others},
	journal={arXiv preprint arXiv:2403.02234},
	year={2024}
}

@article{chen20243dtopia,
	title={3dtopia-xl: Scaling high-quality 3d asset generation via primitive diffusion},
	author={Chen, Zhaoxi and Tang, Jiaxiang and Dong, Yuhao and Cao, Ziang and Hong, Fangzhou and Lan, Yushi and Wang, Tengfei and Xie, Haozhe and Wu, Tong and Saito, Shunsuke and others},
	journal={arXiv preprint arXiv:2409.12957},
	year={2024}
}

@inproceedings{muller2023diffrf,
	title={Diffrf: Rendering-guided 3d radiance field diffusion},
	author={M{\"u}ller, Norman and Siddiqui, Yawar and Porzi, Lorenzo and Bulo, Samuel Rota and Kontschieder, Peter and Nie{\ss}ner, Matthias},
	booktitle={Proceedings of the IEEE/CVF Conference on Computer Vision and Pattern Recognition},
	pages={4328--4338},
	year={2023}
}

@article{cao2023large,
	title={Large-vocabulary 3d diffusion model with transformer},
	author={Cao, Ziang and Hong, Fangzhou and Wu, Tong and Pan, Liang and Liu, Ziwei},
	journal={arXiv preprint arXiv:2309.07920},
	year={2023}
}

@article{cao2024difftf++,
	title={DiffTF++: 3D-aware Diffusion Transformer for Large-Vocabulary 3D Generation},
	author={Cao, Ziang and Hong, Fangzhou and Wu, Tong and Pan, Liang and Liu, Ziwei},
	journal={arXiv preprint arXiv:2405.08055},
	year={2024}
}

@inproceedings{tang2025lgm,
	title={Lgm: Large multi-view gaussian model for high-resolution 3d content creation},
	author={Tang, Jiaxiang and Chen, Zhaoxi and Chen, Xiaokang and Wang, Tengfei and Zeng, Gang and Liu, Ziwei},
	booktitle={European Conference on Computer Vision},
	pages={1--18},
	year={2025},
	organization={Springer}
}

@article{shen2023flexible,
	title={Flexible Isosurface Extraction for Gradient-Based Mesh Optimization.},
	author={Shen, Tianchang and Munkberg, Jacob and Hasselgren, Jon and Yin, Kangxue and Wang, Zian and Chen, Wenzheng and Gojcic, Zan and Fidler, Sanja and Sharp, Nicholas and Gao, Jun},
	journal={ACM Trans. Graph.},
	volume={42},
	number={4},
	pages={37--1},
	year={2023}
}

@inproceedings{lin2023magic3d,
	title={Magic3d: High-resolution text-to-3d content creation},
	author={Lin, Chen-Hsuan and Gao, Jun and Tang, Luming and Takikawa, Towaki and Zeng, Xiaohui and Huang, Xun and Kreis, Karsten and Fidler, Sanja and Liu, Ming-Yu and Lin, Tsung-Yi},
	booktitle={Proceedings of the IEEE/CVF Conference on Computer Vision and Pattern Recognition},
	pages={300--309},
	year={2023}
}

@inproceedings{li2024focaldreamer,
	title={Focaldreamer: Text-driven 3d editing via focal-fusion assembly},
	author={Li, Yuhan and Dou, Yishun and Shi, Yue and Lei, Yu and Chen, Xuanhong and Zhang, Yi and Zhou, Peng and Ni, Bingbing},
	booktitle={Proceedings of the AAAI Conference on Artificial Intelligence},
	volume={38},
	number={4},
	pages={3279--3287},
	year={2024}
}

@article{wang2024prolificdreamer,
	title={Prolificdreamer: High-fidelity and diverse text-to-3d generation with variational score distillation},
	author={Wang, Zhengyi and Lu, Cheng and Wang, Yikai and Bao, Fan and Li, Chongxuan and Su, Hang and Zhu, Jun},
	journal={Advances in Neural Information Processing Systems},
	volume={36},
	year={2024}
}

@article{wu2016learning,
	title={Learning a probabilistic latent space of object shapes via 3d generative-adversarial modeling},
	author={Wu, Jiajun and Zhang, Chengkai and Xue, Tianfan and Freeman, Bill and Tenenbaum, Josh},
	journal={Advances in neural information processing systems},
	volume={29},
	year={2016}
}

@inproceedings{eg3d,
	title={Efficient geometry-aware 3D generative adversarial networks},
	author={Chan, Eric R and Lin, Connor Z and Chan, Matthew A and Nagano, Koki and Pan, Boxiao and De Mello, Shalini and Gallo, Orazio and Guibas, Leonidas J and Tremblay, Jonathan and Khamis, Sameh and others},
	booktitle={Proceedings of the IEEE/CVF Conference on Computer Vision and Pattern Recognition},
	pages={16123--16133},
	year={2022}
}

@article{get3d,
	title={Get3d: A generative model of high quality 3d textured shapes learned from images},
	author={Gao, Jun and Shen, Tianchang and Wang, Zian and Chen, Wenzheng and Yin, Kangxue and Li, Daiqing and Litany, Or and Gojcic, Zan and Fidler, Sanja},
	journal={Advances In Neural Information Processing Systems},
	volume={35},
	pages={31841--31854},
	year={2022}
}

@article{liu2023meshdiffusion,
	title={Meshdiffusion: Score-based generative 3d mesh modeling},
	author={Liu, Zhen and Feng, Yao and Black, Michael J and Nowrouzezahrai, Derek and Paull, Liam and Liu, Weiyang},
	journal={arXiv preprint arXiv:2303.08133},
	year={2023}
}

@article{nam20223d,
	title={3d-ldm: Neural implicit 3d shape generation with latent diffusion models},
	author={Nam, Gimin and Khlifi, Mariem and Rodriguez, Andrew and Tono, Alberto and Zhou, Linqi and Guerrero, Paul},
	journal={arXiv preprint arXiv:2212.00842},
	year={2022}
}

@inproceedings{wang2023rodin,
	title={Rodin: A generative model for sculpting 3d digital avatars using diffusion},
	author={Wang, Tengfei and Zhang, Bo and Zhang, Ting and Gu, Shuyang and Bao, Jianmin and Baltrusaitis, Tadas and Shen, Jingjing and Chen, Dong and Wen, Fang and Chen, Qifeng and others},
	booktitle={Proceedings of the IEEE/CVF Conference on Computer Vision and Pattern Recognition},
	pages={4563--4573},
	year={2023}
}

@inproceedings{shue20233d,
	title={3d neural field generation using triplane diffusion},
	author={Shue, J Ryan and Chan, Eric Ryan and Po, Ryan and Ankner, Zachary and Wu, Jiajun and Wetzstein, Gordon},
	booktitle={Proceedings of the IEEE/CVF Conference on Computer Vision and Pattern Recognition},
	pages={20875--20886},
	year={2023}
}

@article{li2023instant3d,
	title={Instant3d: Fast text-to-3d with sparse-view generation and large reconstruction model},
	author={Li, Jiahao and Tan, Hao and Zhang, Kai and Xu, Zexiang and Luan, Fujun and Xu, Yinghao and Hong, Yicong and Sunkavalli, Kalyan and Shakhnarovich, Greg and Bi, Sai},
	journal={arXiv preprint arXiv:2311.06214},
	year={2023}
}

@article{xu2024instantmesh,
	title={Instantmesh: Efficient 3d mesh generation from a single image with sparse-view large reconstruction models},
	author={Xu, Jiale and Cheng, Weihao and Gao, Yiming and Wang, Xintao and Gao, Shenghua and Shan, Ying},
	journal={arXiv preprint arXiv:2404.07191},
	year={2024}
}

@inproceedings{downs2022google,
	title={Google scanned objects: A high-quality dataset of 3d scanned household items},
	author={Downs, Laura and Francis, Anthony and Koenig, Nate and Kinman, Brandon and Hickman, Ryan and Reymann, Krista and McHugh, Thomas B and Vanhoucke, Vincent},
	booktitle={2022 International Conference on Robotics and Automation (ICRA)},
	pages={2553--2560},
	year={2022},
	organization={IEEE}
}

@inproceedings{wang2025crm,
	title={Crm: Single image to 3d textured mesh with convolutional reconstruction model},
	author={Wang, Zhengyi and Wang, Yikai and Chen, Yifei and Xiang, Chendong and Chen, Shuo and Yu, Dajiang and Li, Chongxuan and Su, Hang and Zhu, Jun},
	booktitle={European Conference on Computer Vision},
	pages={57--74},
	year={2025},
	organization={Springer}
}

@inproceedings{xia2024rgbd,
	title={RGBD Objects in the Wild: Scaling Real-World 3D Object Learning from RGB-D Videos},
	author={Xia, Hongchi and Fu, Yang and Liu, Sifei and Wang, Xiaolong},
	booktitle={Proceedings of the IEEE/CVF Conference on Computer Vision and Pattern Recognition},
	pages={22378--22389},
	year={2024}
}

@article{xiang2024structured,
	title={Structured 3D Latents for Scalable and Versatile 3D Generation},
	author={Xiang, Jianfeng and Lv, Zelong and Xu, Sicheng and Deng, Yu and Wang, Ruicheng and Zhang, Bowen and Chen, Dong and Tong, Xin and Yang, Jiaolong},
	journal={arXiv preprint arXiv:2412.01506},
	year={2024}
}

@inproceedings{lim2017enhanced,
	title={Enhanced deep residual networks for single image super-resolution},
	author={Lim, Bee and Son, Sanghyun and Kim, Heewon and Nah, Seungjun and Mu Lee, Kyoung},
	booktitle={Proceedings of the IEEE conference on computer vision and pattern recognition workshops},
	pages={136--144},
	year={2017}
}

@article{tang2023dreamgaussian,
	title        = {DreamGaussian: Generative Gaussian Splatting for Efficient 3D Content Creation},
	author       = {Tang, Jiaxiang and Ren, Jiawei and Zhou, Hang and Liu, Ziwei and Zeng, Gang},
	year         = 2023,
	journal      = {arXiv preprint arXiv:2309.16653}
}

@article{shi2023mvdream,
	title        = {MVDream: Multi-view Diffusion for 3D Generation},
	author       = {Shi, Yichun and Wang, Peng and Ye, Jianglong and Long, Mai and Li, Kejie and Yang, Xiao},
	year         = 2023,
	journal      = {arXiv preprint arXiv:2308.16512}
}

@article{wang2023imagedream,
	title        = {ImageDream: Image-Prompt Multi-view Diffusion for 3D Generation},
	author       = {Wang, Peng and Shi, Yichun},
	year         = 2023,
	journal      = {arXiv preprint arXiv:2312.02201}
}

@misc{openlrm,
	title        = {OpenLRM: Open-Source Large Reconstruction Models},
	author       = {Zexin He and Tengfei Wang},
	year         = 2023,
	howpublished = {\url{https://github.com/3DTopia/OpenLRM}}
}

@article{xu2024grm,
	title        = {Grm: Large gaussian reconstruction model for efficient 3d reconstruction and generation},
	author       = {Xu, Yinghao and Shi, Zifan and Yifan, Wang and Chen, Hansheng and Yang, Ceyuan and Peng, Sida and Shen, Yujun and Wetzstein, Gordon},
	year         = 2024,
	journal      = {arXiv preprint arXiv:2403.14621}
}

@article{zhang2024geolrm,
	title        = {GeoLRM: Geometry-Aware Large Reconstruction Model for High-Quality 3D Gaussian Generation},
	author       = {Zhang, Chubin and Song, Hongliang and Wei, Yi and Chen, Yu and Lu, Jiwen and Tang, Yansong},
	year         = 2024,
	journal      = {arXiv preprint arXiv:2406.15333}
}

@article{wei2024meshlrm,
	title        = {Meshlrm: Large reconstruction model for high-quality mesh},
	author       = {Wei, Xinyue and Zhang, Kai and Bi, Sai and Tan, Hao and Luan, Fujun and Deschaintre, Valentin and Sunkavalli, Kalyan and Su, Hao and Xu, Zexiang},
	year         = 2024,
	journal      = {arXiv preprint arXiv:2404.12385}
}

@inproceedings{
	liu2024meshformer,
	title={MeshFormer : High-Quality Mesh Generation with 3D-Guided Reconstruction Model},
	author={Minghua Liu and Chong Zeng and Xinyue Wei and Ruoxi Shi and Linghao Chen and Chao Xu and Mengqi Zhang and Zhaoning Wang and Xiaoshuai Zhang and Isabella Liu and Hongzhi Wu and Hao Su},
	booktitle={The Thirty-eighth Annual Conference on Neural Information Processing Systems},
	year={2024},
	url={https://openreview.net/forum?id=x7pjdDod6Z}
}

@article{jun2023shap,
	title        = {Shap-e: Generating conditional 3d implicit functions},
	author       = {Jun, Heewoo and Nichol, Alex},
	year         = 2023,
	journal      = {arXiv preprint arXiv:2305.02463}
}

@inproceedings{NIPS2017pointnet,
	author = {Qi, Charles Ruizhongtai and Yi, Li and Su, Hao and Guibas, Leonidas J},
	title = {PointNet++: Deep Hierarchical Feature Learning on Point Sets in a Metric Space},
	volume = {30},
	year = {2017}
}

@inproceedings{lan2024ga,
	title={GaussianAnything: Interactive Point Cloud Latent Diffusion for 3D Generation},
	author={Lan, Yushi and Zhou, Shangchen and Lyu, Zhaoyang and Hong, Fangzhou and Yang, Shuai and Dai, Bo and Pan, Xingang and Loy, Chen Change},
	year={2025},
	booktitle={ICLR}
}

@inproceedings{chen2024sar3d,
	title={SAR3D: Autoregressive 3D Object Generation and Understanding via Multi-scale 3D VQVAE},
	author={Chen, Yongwei and Lan, Yushi and Zhou, Shangchen and Wang, Tengfei and Pan, Xingang},
	booktitle={CVPR},
	year={2025}
}

@inproceedings{lan2024ln3diff,
	title={LN3Diff: Scalable Latent Neural Fields Diffusion for Speedy 3D Generation}, 
	author={Lan, Yushi and Hong, Fangzhou and Yang, Shuai and Zhou, Shangchen and Meng, Xuyi and Dai, Bo and Pan, Xingang and Loy, Chen Change},
	year={2024},
	booktitle={ECCV},
}

@article{luo20243denhancer,
	title={3DEnhancer: Consistent Multi-View Diffusion for 3D Enhancement}, 
	author={Yihang Luo and Shangchen Zhou and Yushi Lan and Xingang Pan and Chen Change Loy},
	booktitle={arXiv preprint arXiv:2412.18565},
	year={2024},
}

@article{cao2025physx,
	title={PhysX: Physical-Grounded 3D Asset Generation},
	author={Cao, Ziang and Chen, Zhaoxi and Pan, Liang and Liu, Ziwei},
	journal={arXiv preprint arXiv:2507.12465},
	year={2025}
}

@article{cao2025physxanything,
	title={PhysX-Anything: Simulation-Ready Physical 3D Assets from Single Image},
	author={Cao, Ziang and Hong, Fangzhou and Chen, Zhaoxi and Pan, Liang and Liu, Ziwei},
	journal={arXiv preprint arXiv:2511.13648},
	year={2025}
}

@article{wu2025qwen,
	title={Qwen-image technical report},
	author={Wu, Chenfei and Li, Jiahao and Zhou, Jingren and Lin, Junyang and Gao, Kaiyuan and Yan, Kun and Yin, Sheng-ming and Bai, Shuai and Xu, Xiao and Chen, Yilei and others},
	journal={arXiv preprint arXiv:2508.02324},
	year={2025}
}

 \vspace{-40pt}
	\begin{IEEEbiography}[{\includegraphics[width=1in,height=1.25in,clip,keepaspectratio]{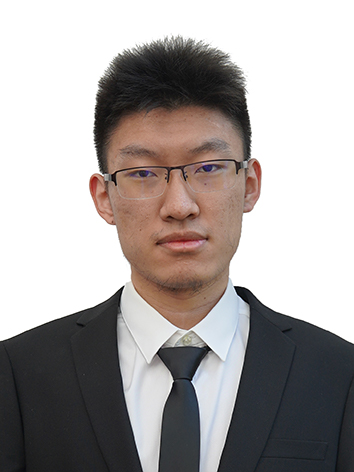}}]{Ziang Cao}
		is currently pursuing a Ph.D. in the College of Computing and Data Science at Nanyang Technological University, supervised by Prof. Ziwei Liu. His research interests lie on computer vision, deep learning, and 3D generation.
	\end{IEEEbiography}
    
 \vspace{-40pt}
 
	\begin{IEEEbiography}[{\includegraphics[width=1in,height=1.25in,clip,keepaspectratio]{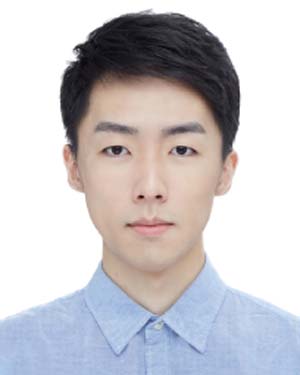}}]{Zhaoxi Chen}
		 received the bachelor’s
degree from Tsinghua University, in 2021. He
is currently working toward the PhD degree with
MMLabatNTU, Nanyang Technological University,
supervised by Prof. Ziwei Liu. He received the AISG
PhD Fellowship in 2021. His research interests include
inverse rendering and 3D generative models.
He has published several papers in CVPR, ICCV,
ECCV, ICLR, IEEE Transactions on Pattern Analysis
and Machine Intelligence, and ACM Transactions on
Graphics. He also served as a reviewer for CVPR,
ICCV, NeurIPS, ACM Transactions on Graphics, and International Journal of
Computer Vision

	\end{IEEEbiography}

     \vspace{-40pt}
	
	\begin{IEEEbiography}[{\includegraphics[width=1in,height=1.25in,clip,keepaspectratio]{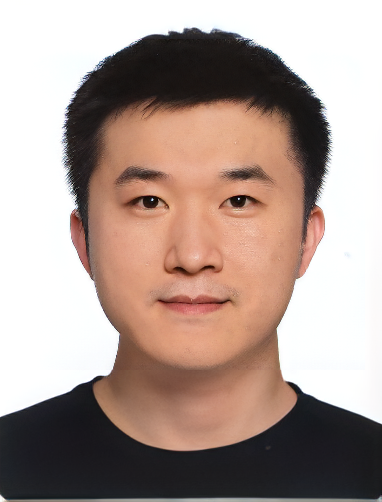}}]{Liang Pan}
		is presently a Researcher at the Shanghai AI Laboratory. He earned his Ph.D. in Mechanical Engineering from the National University of Singapore (NUS) in 2019. He then served as a Research Fellow at the S-Lab of Nanyang Technological University from 2020 to 2023. His research focuses on computer vision, 3D point clouds, and virtual humans. He has made top-tier publications in relevant conferences and journals. Furthermore, he actively contributes to the academic community by serving as a reviewer for esteemed conferences and journals in computer vision, machine learning, and robotics.
	\end{IEEEbiography}
	
	
	
	 \vspace{-40pt}
	
	\begin{IEEEbiography}[{\includegraphics[width=1in,height=1.25in,clip,keepaspectratio]{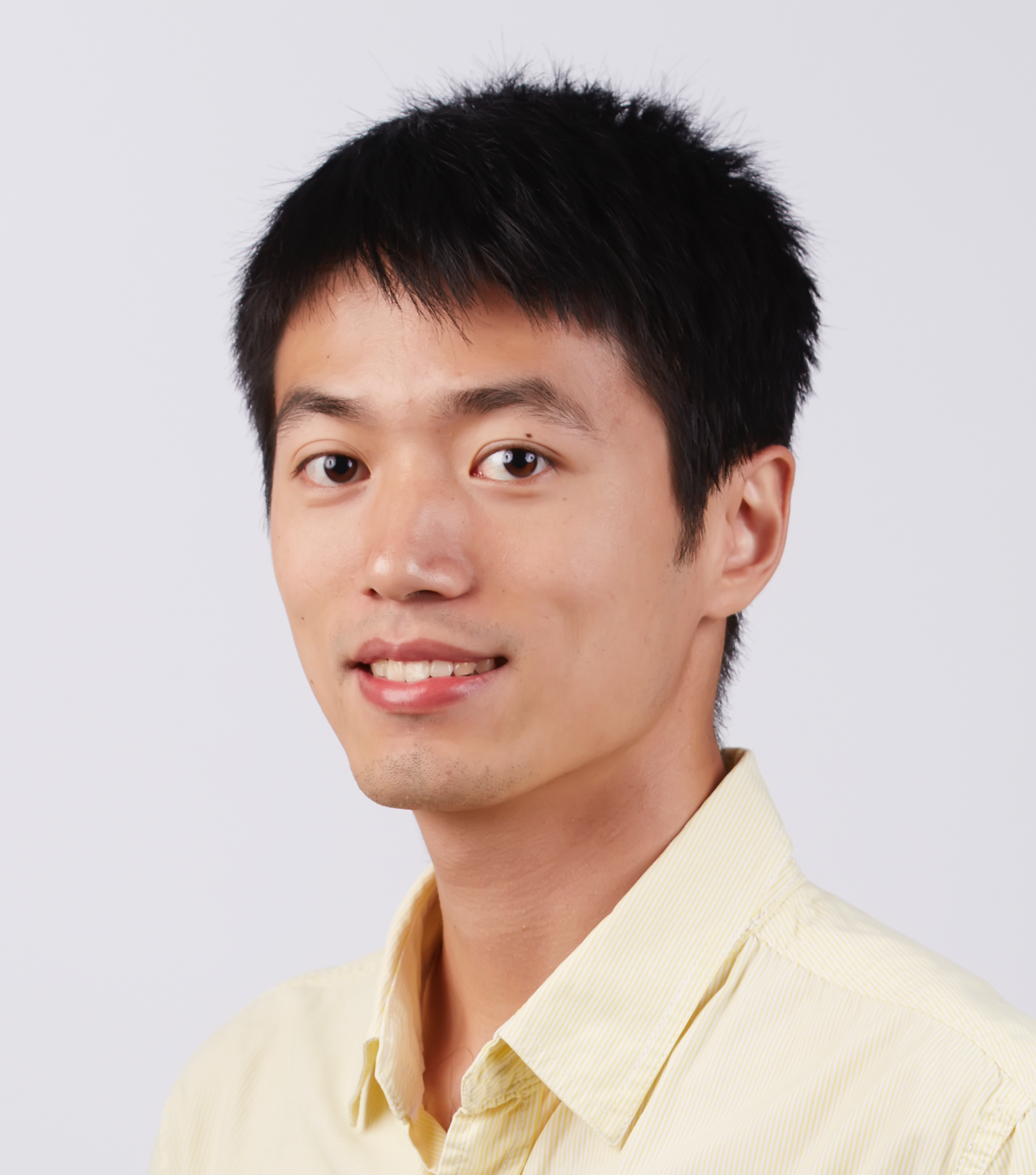}}]{Ziwei Liu}
		Ziwei Liu is currently an Assistant Professor at Nanyang Technological University (NTU). Previously, he was a senior research fellow at the Chinese University of Hong Kong and a postdoctoral researcher at the University of California, Berkeley. Ziwei received his Ph.D. from the Chinese University of Hong Kong in 2017. His research revolves around computer vision/graphics, machine learning, and robotics. He has published extensively on top-tier conferences and journals in relevant fields, including CVPR, ICCV, ECCV, NeurIPS, IROS, SIGGRAPH, TOG, and TPAMI. He is the recipient of the Microsoft Young Fellowship, Hong Kong PhD Fellowship, ICCV Young Researcher Award, and HKSTP best paper award.
		
	\end{IEEEbiography}

\end{document}